%% file: arxiv.tex
\pdfoutput=1
\documentclass{article}

\usepackage{iclr2027_conference,times}
\iclrfinalcopy
\input{preamble}

\newcommand{\parag}[1]{\vspace{-3mm}\paragraph{#1}}

\providecommand{\figorbox}[2]{\IfFileExists{#1}{\includegraphics[width=\linewidth]{#1}}{\fbox{\parbox[c][#2][c]{0.96\linewidth}{\centering\small Placeholder: \texttt{\detokenize{#1}}}}}}

\title{PhysPlan: Grounded Physical State Reasoning and Graph-Guided Optimization for Physically Plausible Video Generation}

\author{
Minh-Loi Nguyen$^{1,2,\dagger}$ \quad
Xuan-Vu Le$^{1,2,\dagger}$ \quad
Trung-Nghia Le$^{1,2,\ddagger,*}$ \quad
Tam V. Nguyen$^{3}$ \\ 
\textbf{Minh-Triet Tran}$^{1,2}$ \quad \textbf{Thanh-Toan Do}$^{4,\ddagger}$\\
$^1$University of Science, Ho Chi Minh City, Vietnam\\
$^2$Vietnam National University, Ho Chi Minh City, Vietnam\\
$^3$University of Dayton, Dayton, Ohio, USA\\
$^4$Monash University, Melbourne, Victoria, Australia\\
\texttt{\{22120189,22120438\}@student.hcmus.edu.vn}\\
\texttt{\{tmtriet,ltnghia\}@fit.hcmus.edu.vn}\\
\texttt{tamnguyen@udayton.edu,toan.do@monash.edu}\\
$^\dagger$Equal contribution. $^\ddagger$Co-supervising. $^*$Corresponding author.
}

\begin{document}

\maketitle

\begin{center}
\begin{minipage}{\textwidth}
    \centering
    \vspace{-5mm}
    \figorbox{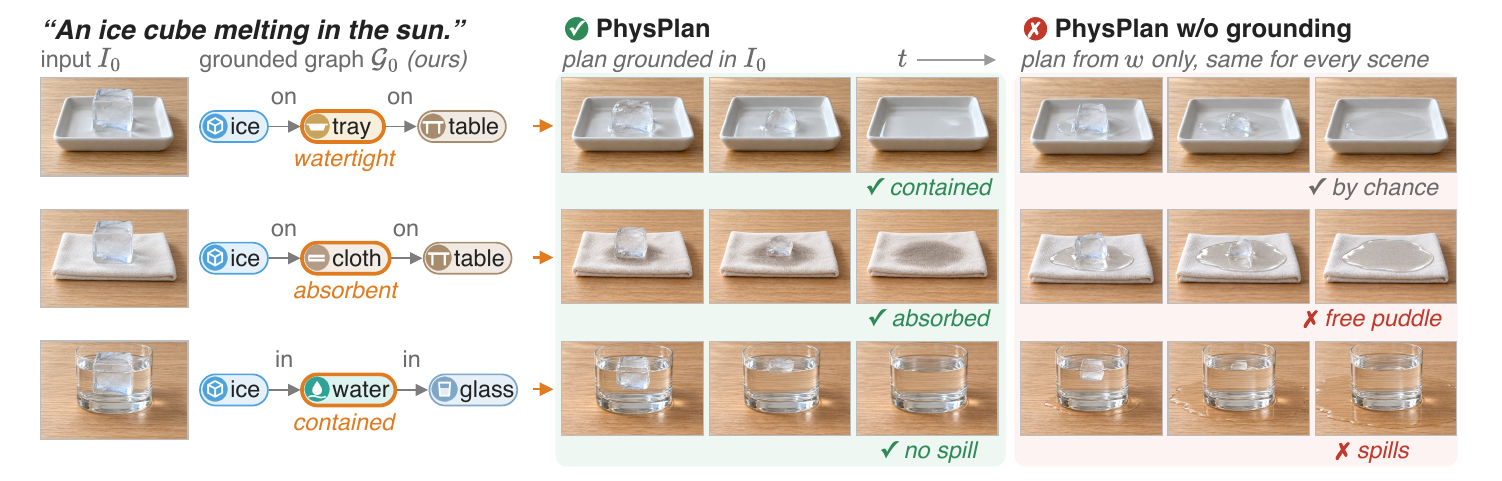}{4.4cm}
    \vspace{-5mm}
    \captionof{figure}{\textbf{Same words, different scenes.} ``An ice cube melting in the sun''
    says what occurs; the scene decides where the meltwater goes: it stays in a watertight
    tray, soaks into an absorbent cloth, and joins the water in a glass. PhysPlan plans on a
    state graph grounded in the input frame. Without grounding, the plan comes from the
    prompt alone and yields the same free puddle in every scene.}
    \label{fig:teaser}
\end{minipage}
\end{center}
\vspace{-2mm}

\input{sections/0.abstract}
\input{sections/1.introduction}
\input{sections/2.relatedworks}
\input{sections/3.method}
\input{sections/experiment_v2}
\input{sections/6.conclusion}

\section*{Acknowledgment} 

We sincerely thank Prof. Trung Le (Monash University) for his valuable discussions, insightful comments, and helpful suggestions.

\section*{AI Use Statement}

LLMs/VLMs were used as components of the proposed system and experimental pipeline. Specifically, Gemini 3 Flash was used for grounded physical state reasoning, including scene parsing, phenomenon decomposition, graph-edit generation, and event timing estimation; Gemini 3 Pro Image was used to render keyframes from the graph-derived editing instructions; and the Gemini API was used to generate initial frames for the PhyGenBench evaluation. 

ChatGPT was used solely for grammar correction and language polishing. All technical methodology, experiments, and results were developed, conducted, and verified by the authors.

\subsubsection*{Reproducibility Statement}
Implementation details, exact VLM instructions, optimization pseudocode, evaluation protocols, and additional qualitative results are provided in the supplementary material. The complete source code and generated video samples will be released upon publication.

\subsubsection*{Ethics Statement}
PhysPlan can improve the realism of generated videos and may therefore be misused to create deceptive synthetic media. We advocate provenance mechanisms such as digital watermarking and restrict the intended use of this work to creative, research, and educational applications. The de-identified human evaluation used adult participants who provided informed consent; no personally identifiable information was collected.

\bibliography{main}
\bibliographystyle{iclr2027_conference}

\newpage
\appendix

\input{sections/7.appendix}

\end{document}

%% file: preamble.tex
\usepackage[T1]{fontenc}
\usepackage{graphicx}
\usepackage{booktabs}
\usepackage{amsmath}
\usepackage{amssymb}
\usepackage{multirow}
\usepackage{wrapfig}
\usepackage{needspace}
\usepackage{algorithm}
\usepackage{algpseudocode}
\usepackage{enumitem}
\usepackage{listings}
\usepackage[table]{xcolor}
\usepackage{tabularx}
\usepackage{siunitx}
\usepackage{tcolorbox}
\usepackage{xspace}
\usepackage{url}
\usepackage{soul}
\usepackage{capt-of}

\usepackage{microtype}

\usepackage[labelfont=bf]{caption}
\setlist{nosep}

\definecolor{ppblue}{HTML}{1F5FA8}   %
\definecolor{ppred}{HTML}{B03A2E}    %
\definecolor{ppurl}{HTML}{7A3E9D}    %
\definecolor{oursrow}{HTML}{E8F1FB}  %
\definecolor{ppgain}{HTML}{2E7D32}   %
\definecolor{ppdrop}{HTML}{B03A2E}   %
\definecolor{lightgray}{gray}{0.9}   %
\definecolor{promptvar}{HTML}{008080}
\definecolor{pptoc}{HTML}{1F3A68}    %
\sethlcolor{yellow}

\usepackage{hyperref}
\hypersetup{
  colorlinks   = true,
  citecolor    = ppblue,
  linkcolor    = ppred,
  urlcolor     = ppurl,
  bookmarksnumbered = true,
  pdftitle     = {PhysPlan: Grounded Physical State Reasoning and Graph-Guided Optimization for Physically Plausible Video Generation},
  pdfkeywords  = {video generation, physical plausibility, scene graphs,
                  training-free guidance, vision-language models},
}
\usepackage[nameinlink,noabbrev,capitalise]{cleveref}
\crefname{equation}{Eq.}{Eqs.}
\Crefname{equation}{Equation}{Equations}
\crefname{algorithm}{Algorithm}{Algorithms}
\crefname{appendix}{Appendix}{Appendices}
\creflabelformat{equation}{#2(#1)#3}

\newcommand{\eg}{\emph{e.g.}\xspace}
\newcommand{\ie}{\emph{i.e.}\xspace}

\newcommand{\best}[1]{\textbf{#1}}
\newcommand{\second}[1]{\underline{#1}}
\newcommand{\gain}[1]{\textsubscript{\,\color{ppgain}\scriptsize #1}}
\newcommand{\drop}[1]{\textsubscript{\,\color{ppdrop}\scriptsize #1}}

\lstdefinelanguage{json}{
  numbers=none,
  breaklines=true,
  frame=none,
  tabsize=2,
  stringstyle=\color{black},
  showstringspaces=false,
  identifierstyle=\color{black},
  keywords={false,true},
  keywordstyle=\color{blue}
}

\newtcolorbox{promptbox}{
  colback=gray!5,
  colframe=black!80,
  boxrule=0.5pt,
  arc=4pt,
  left=10pt,
  right=10pt,
  top=10pt,
  bottom=10pt,
  fontupper=\small
}

\newcommand{\tocleaders}{\leaders\hbox to 0.9em{\hss.\hss}\hfill}
\newcommand{\appsec}[1]{%
  \par\addvspace{13pt}\noindent
  \hyperref[#1]{\textcolor{pptoc}{\textbf{(\ref*{#1})~\nameref*{#1}}}}%
  \nobreak\tocleaders\nobreak\hyperref[#1]{\textcolor{pptoc}{\textbf{\pageref*{#1}}}}\par}
\newcommand{\appsub}[1]{%
  \par\addvspace{6pt}\noindent\hspace*{1.8em}%
  \hyperref[#1]{\textcolor{pptoc}{(\ref*{#1})~\nameref*{#1}}}%
  \nobreak\tocleaders\nobreak\hyperref[#1]{\textcolor{pptoc}{\pageref*{#1}}}\par}

%% file: sections/0.abstract.tex
\begin{abstract}
Video diffusion models (VDMs) synthesize photorealistic content, yet they often fail to
follow the course that a physical phenomenon should take within a given scene. Recent
training-free methods let a vision-language model (VLM) plan the phenomenon and guide a
frozen VDM toward the plan; however, such plans are derived from the prompt and consumed as
whole keyframes or trajectories, which leaves unspecified where the consequences land in
the observed scene and turns incidental visual details into optimization targets. We
observe that a phenomenon specified in words unfolds as sparse, local changes to the
physical state of the observed scene. Building on this observation, we present
\textbf{PhysPlan}, a training-free image-to-video framework that represents a phenomenon as
a grounded state graph and uses this graph to decide what, where, and when the guidance
constrains. \emph{Grounded Physical State Reasoning} decomposes the phenomenon into
physical deltas, each stating which objects change, to what state, and by which physical rule,
and translates each delta into graph edits, verified by
deterministic checks, that leave all other objects unchanged. \emph{Graph-Guided Test-Time Optimization} renders a
keyframe for each state, measures the denoised estimates only along the properties selected
by the edits, and concentrates the update on the edited objects. On PhyGenBench and Physics-IQ, PhysPlan raises its base model from 0.52 to 0.77 and from
27.1 to 38.2, surpassing the strongest prior I2V method (0.60 and 34.6), and lowers FVD by
over 20\%.
\par\noindent\textbf{Project page}:~\href{https://physplan.github.io}{{\setlength{\fboxsep}{1.5pt}\colorbox[HTML]{FDF0F4}{\strut\textcolor[HTML]{D6336C}{\texttt{physplan.github.io}}}}}.
\end{abstract}

%% file: sections/1.introduction.tex
\section{Introduction}
\label{sec:intro}

Video diffusion models (VDMs) \citep{blattmann2023stable,yang2025cogvideox,hacohen2024ltx}
synthesize photorealistic and temporally coherent videos, drawing attention to their
potential as world simulators \citep{brooks2024video}. Yet VDMs mainly reproduce statistical
patterns of their training data rather than the physical processes behind them
\citep{kang2025how}: they often fail to follow the course that a phenomenon should take,
\eg, an ice cube vanishes without leaving water behind, or uninvolved objects change their
appearance \citep{motamed2025generative,meng2024towards}.

A growing line of work therefore decouples physical reasoning from visual synthesis: a
language or vision-language model (VLM) plans the phenomenon as trajectories
\citep{yang2025vlipp}, keyframes \citep{huang2025vchain,phyrpr2026}, both
\citep{causalmotion2026}, or chains of events over scene graphs \citep{wang2026physically},
and the plan is injected into a frozen VDM at inference time. Two gaps remain.
\emph{First, the plan is derived from the prompt, whereas the outcome is determined by the
scene.} Where the consequences of a phenomenon land depends on the physical relations among
the observed objects (\cref{fig:teaser}). A prompt-derived plan does not know which passive
objects the scene contains, and when states are described anew as complete scenes, nothing
ties one state to the next, so the VLM may omit passive objects, alter uninvolved ones, or
lose object identities. \emph{Second, the plan is consumed as whole frames.} A keyframe
commits to details the plan never intended, such as the exact contour of a puddle; guiding
toward whole keyframes \citep{jang2025frame} turns these details into targets, and the
gradient disturbs content that should remain unchanged.

We observe that a phenomenon unfolds as \emph{sparse, local changes to the physical state
of the observed scene}: a few objects and relations change at a time, while the rest
persists. We therefore represent the phenomenon as changes to a state graph grounded in the
observed frame, and let the same graph decide what the guidance constrains. Based on this
idea, we propose \textbf{PhysPlan}, a general training-free image-to-video framework with
two modules. \emph{Grounded Physical State Reasoning} (GPSR) parses the observed frame into a
state graph and decomposes the phenomenon into causally ordered \emph{physical deltas}, each
stating the new state of every changing object, the qualitative physical rule that produces
it, and the share of the video the event takes. The same VLM translates each delta into
typed graph edits, verified by deterministic checks, so every object outside the edits is
preserved by construction. \emph{Graph-Guided Test-Time Optimization} (GTO) renders a
keyframe for each state and guides a frozen VDM by measuring its denoised estimates at the
anchor frames only along the properties selected by the edits, with the update concentrated
on the edited objects. Since the modules communicate only through the state graph, all
models are used off the shelf and can be replaced. Our contributions are:
\begin{itemize}[leftmargin=2em]
    \item PhysPlan, a general training-free framework in which a state graph grounded in the
    observed frame determines what, where, and when a frozen VDM is guided.
    \item Grounded Physical State Reasoning, which evolves the state graph only through
    physical deltas realized as verified edits, making every change traceable and every
    unchanged object explicit.
    \item Graph-Guided Test-Time Optimization, which selects the measured properties by the
    edit types, compares objects rather than pixels, and weights the update by the regions
    of the edited objects.
    \item On PhyGenBench \citep{meng2024towards} and Physics-IQ \citep{motamed2025generative},
    PhysPlan raises its base model from 0.52 to 0.77 and from 27.1 to 38.2 while improving
    FVD, VBench quality, and human preference; ablations confirm that both gaps matter.
\end{itemize}

%% file: sections/2.relatedworks.tex
\section{Related Work}
\label{sec:related}

\parag{Physically plausible video generation.}
Large-scale VDMs \citep{blattmann2023stable,yang2025cogvideox,hacohen2024ltx,zheng2024open,wan2025wan}
achieve high visual fidelity but often violate physical laws
\citep{meng2024towards,motamed2025generative}, and scaling alone does not resolve this gap
\citep{kang2025how}. Simulator-based methods \citep{liu2024physgen} model exact dynamics but
are limited to supported phenomena, while training-based methods
\citep{chen2026hierarchical,cai2026phygdpo} require curated data and backbone-specific
retraining. Prompt-based methods \citep{xue2025phyt2v} add physical details to prompts but
remain ambiguous about scene-specific outcomes. Inference-time methods instead suppress
implausible dynamics with counterfactual prompts \citep{reasonimplausibility2025} or refine
samples using the generator's scores \citep{proprio2026}. PhysPlan is training- and
simulator-free, while imposing positive, object-level constraints grounded in the observed scene.

\parag{Reasoning-guided video generation.}
Early planners predict layouts for images \citep{feng2023layoutgpt} and videos
\citep{lian2024llmgrounded,lin2023videodirectorgpt}, but not physical state changes.
Recent methods plan physical phenomena as trajectories \citep{yang2025vlipp}, keyframes
\citep{huang2025vchain,phyrpr2026}, both \citep{causalmotion2026}, or events over evolving
scene graphs \citep{wang2026physically}. These plans are derived from the prompt rather than
the observed scene and are typically consumed as complete scenes; even
\citet{wang2026physically} conditions the VDM on whole keyframes. PhysPlan instead grounds
the graph in the observed frame and evolves it through verified edits, following the frame
problem \citep{mccarthy1969frame}. PhysPlan relies on VLMs for what they do well: qualitative physical reasoning, such as
judging which objects a phenomenon affects, how their states change, and in what order.
Their quantitative estimates remain unreliable \citep{quantiphy2025}, so PhysPlan never asks
for them.

\parag{Training-free guidance of diffusion models.}
Loss-based guidance steers pretrained diffusion models by differentiating objectives on
estimated clean samples \citep{bansal2023universal,yu2023freedom,shen2024understanding}.
Frame Guidance \citep{jang2025frame} extends this to large VDMs by decoding latent slices
around guided frames, but matches each target as a whole, e.g., with an L2 keyframe loss,
making unspecified details optimization targets and potentially altering unchanged content.
PhysPlan instead compares anchors with keyframes at the object level and only on properties
selected by graph edits, while restricting updates to the corresponding edited regions.

%% file: sections/3.method.tex
\section{Proposed Method}
\label{sec:method}

\providecommand{\opname}[1]{\textsc{#1}}

\begin{figure*}[t]
    \centering
    \figorbox{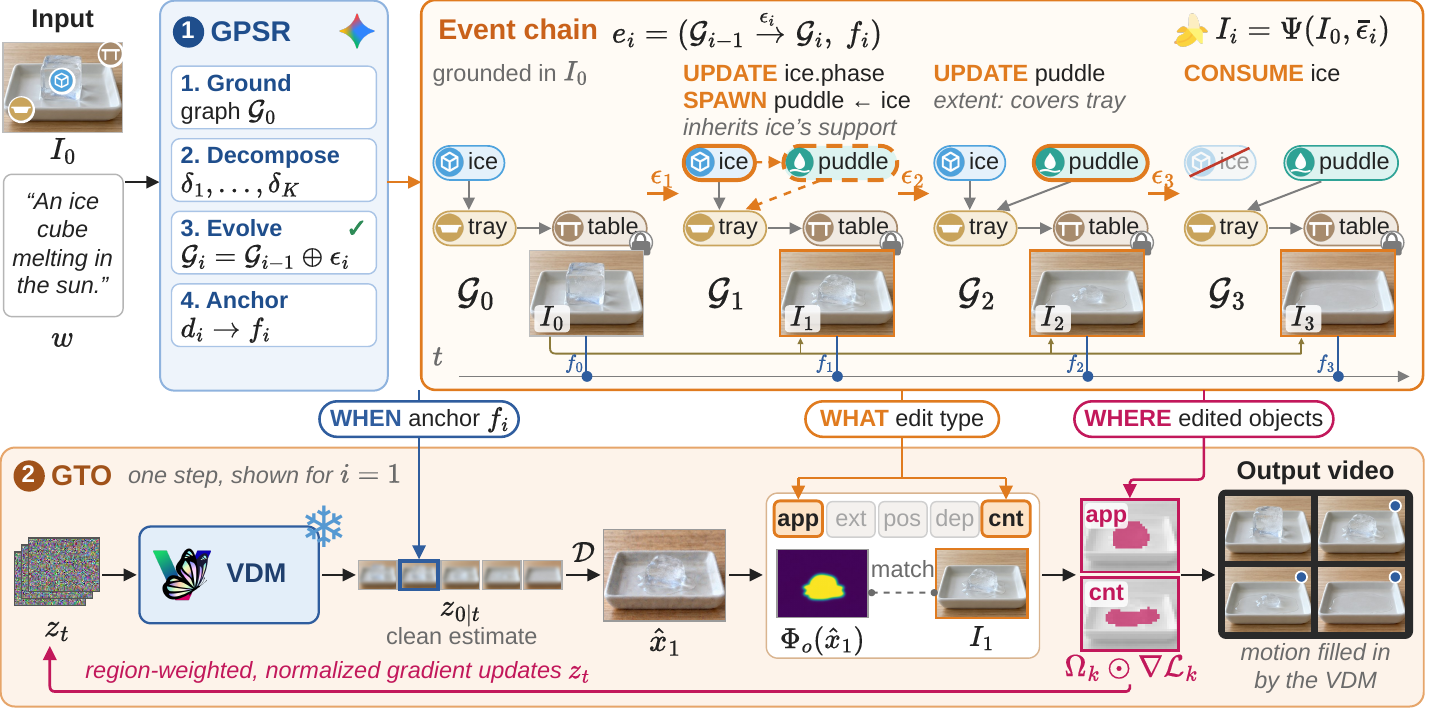}{6.5cm}
    \vspace{-2mm}
    \caption{\textbf{Overview of PhysPlan.} \textbf{Top:} GPSR (\cref{sec:stage1}) grounds
    $I_0$ and $w$ as a state graph and evolves it through verified edits $\epsilon_i$ into
    an event chain, anchoring each state $\mathcal{G}_i$ to a frame $f_i$; unedited objects
    are carried over (lock). Each keyframe $I_i$ is rendered from $I_0$ and the net edits
    $\bar{\epsilon}_i$. \textbf{Bottom:} GTO (\cref{sec:stage2}), shown for $i=1$: the anchor
    selects \emph{when} the clean estimate of the frozen VDM is compared with $I_i$, the edit
    type selects \emph{what} is measured, and the edited objects define \emph{where} the
    gradient acts. The VDM fills in the motion between anchors. Icons: Gemini~3 Flash (VLM),
    Gemini~3 Pro Image ($\Psi$), CogVideoX-I2V-5B (VDM).}
    \label{fig:overview}
    \vspace{-3mm}
\end{figure*}

Given an observed frame $I_0$ and a text prompt $w$ describing a physical phenomenon, our
goal is to generate a video of $F$ frames that starts from $I_0$ and is physically
plausible. This involves deciding what should happen in the scene and making the VDM
generate it. PhysPlan addresses these with two modules (\cref{fig:overview}): \emph{Grounded
Physical State Reasoning} (GPSR, \cref{sec:stage1}) predicts how the scene should change, and
\emph{Graph-Guided Test-Time Optimization} (GTO, \cref{sec:stage2}) guides a frozen VDM to
follow this prediction. The two modules are connected by an \emph{event chain}
\begin{equation}
    \mathcal{G}_0 \xrightarrow{e_1} \mathcal{G}_1 \xrightarrow{e_2} \cdots
    \xrightarrow{e_K} \mathcal{G}_K,
    \qquad
    e_i = \bigl(\mathcal{G}_{i-1} \xrightarrow{\epsilon_i} \mathcal{G}_i,\; f_i\bigr),
    \label{eq:interface}
\end{equation}
where $\mathcal{G}_i$ is a state graph, $\epsilon_i$ is the graph edits of the $i$-th event,
and $f_i$ is the frame at which $\mathcal{G}_i$ should hold. Thus, each event specifies what to measure through its edit type, where through its edited objects, and when through its anchor; the VDM generates the remaining content and motion.

\subsection{Grounded Physical State Reasoning}
\label{sec:stage1}

\begin{figure}[t]
    \centering
    \figorbox{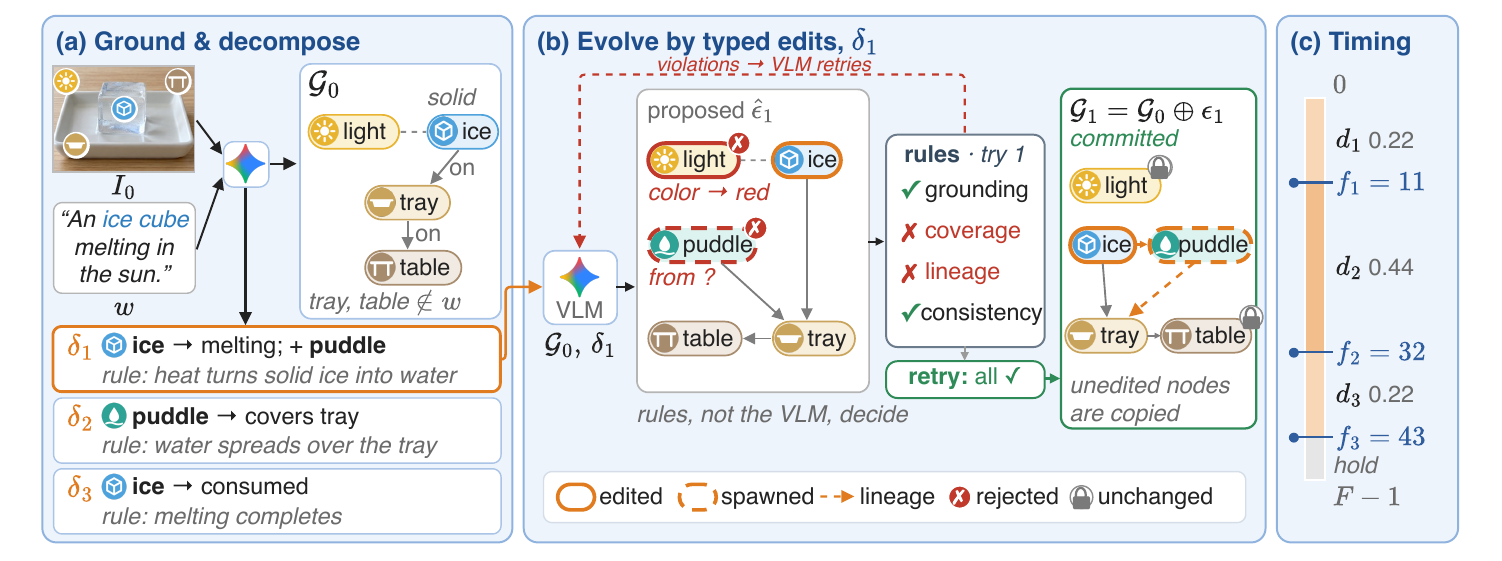}{4.8cm}
    \vspace{-5mm}
    \caption{\textbf{Grounded Physical State Reasoning.} (a) The VLM grounds the scene as
    $\mathcal{G}_0$, including objects not mentioned in $w$, and decomposes the phenomenon
    into deltas with qualitative rules. (b) Each delta is translated into typed edits.
    Deterministic checks reject invalid proposals, here an edit to an object the delta does
    not name (coverage) and a new object without a source (lineage); the VLM retries, and
    only a passing proposal is committed, with all other nodes copied. (c) The share $d_i$
    of each event places its state at an anchor frame $f_i$.}
    \label{fig:stage1}
    \vspace{-3mm}
\end{figure}

The effect of a phenomenon depends on the physical relations in the scene: melting ice wets
the tray it rests on, but not the table beside it (\cref{fig:teaser}), and only a few objects
change at a time. We therefore represent the scene as a state graph grounded in $I_0$ and the
phenomenon as a sequence of small edits to it (\cref{fig:stage1}). Since VLMs are unreliable
at physical quantitative reasoning \citep{physbench2025,causalphys2026,quantiphy2025}, the
edits are qualitative and verified by deterministic checks, and the VLM never regenerates a
full state, so objects that should not change stay unchanged.

\parag{Grounded Phenomenon Decomposition.}
A VLM parses $I_0$ and $w$ into a state graph $\mathcal{G}_0=(\mathcal{V}_0,\mathcal{E}_0)$
that records the state of each object and how objects relate. \emph{Nodes} are all objects
visible in $I_0$, not only those named in $w$: a prompt about melting ice never mentions the
tray, yet the water spreads onto it. Each node has a persistent identifier, a category, and
attributes under six fixed keys with open-vocabulary values: phase, integrity (\eg, cracked),
surface (\eg, wet), color, extent (size), and configuration (pose or shape, \eg, tipped over).
Fixed keys make states comparable across graphs; open values describe phenomenon-specific
states. \emph{Edges} $(a,r,b)$ take $r$ from a fixed set, since each relation type is checked
and guided differently: physical relations (support, contact, containment, attachment) carry
effects between objects, and spatial relations (left of, above, in front of, near) fix the
layout. Relations observable in the image plane are \emph{planar} and guided by
$\mathcal{L}_{\mathrm{pos}}$; containment and in front of are \emph{depth} relations, guided
by $\mathcal{L}_{\mathrm{depth}}$ (\cref{sec:stage2}).

The VLM then decomposes the phenomenon into a causally ordered sequence of physical deltas
$\Delta = (\delta_1, \dots, \delta_K)$. Each delta $\delta_i = (\mathcal{S}_i, d_i)$ describes
the state of the scene at the end of one event: $d_i \in (0, 1)$ is the fraction of the video
that the event takes, and each entry $(o, s_o, r_o) \in \mathcal{S}_i$ gives a changing
object $o$, its new state $s_o$, and the physical rule $r_o$ that produces it. Rules are short qualitative statements without numbers: the VLM judges which objects change, how, and in what order, rather than estimating quantities
\citep{quantiphy2025}; requiring a rule for every state makes each change follow from a cause.
For example, melting ice gives (ice\#1, \emph{partially melted}, \emph{ice above its melting
point turns into water}) and (tray\#2, \emph{wet}, \emph{liquid flows down onto its
support}). A delta covers all objects that change in the event, since unrelated objects can
change at the same time (\eg, ice and butter on separate plates). Objects are referred to by
node identifiers, new objects get identifiers linked to their source, and unlisted
objects stay unchanged. The whole sequence is predicted in one pass.

\parag{Delta-driven Graph Evolution.}
As text, the deltas cannot be checked or measured, and letting the VLM rewrite the full graph
would also change objects that the delta does not mention. Instead, the same VLM translates
each delta into typed edits $\epsilon_i = \mathrm{VLM}(\mathcal{S}_i, \mathcal{G}_{i-1})$ from
the operator set $\mathcal{O}$ (\cref{tab:operators}), and the next graph is
$\mathcal{G}_i = \mathcal{G}_{i-1} \oplus \epsilon_i$, where $\oplus$ applies the edits. For example, (tray\#2, \emph{wet}) becomes
\opname{Update}(tray\#2, surface, wet), and ice turning into a puddle becomes
\opname{Spawn}(puddle\#3 $\leftarrow$ ice\#1) and \opname{Consume}(ice\#1). The edits are
verified by four deterministic checks: (i) \emph{grounding}, edits use existing or newly
spawned nodes and valid keys or relation types; (ii) \emph{coverage}, exactly the objects in
$\mathcal{S}_i$ are edited; (iii) \emph{lineage}, nodes are added only by \opname{Spawn} and
removed only by \opname{Consume}; and (iv) \emph{consistency}, no attribute is set twice, and
support and containment stay acyclic. Violations are returned to the VLM, which retries.
Unedited nodes are copied unchanged, so every change comes from a delta. Each edit type also sets how its change is measured (\cref{tab:operators}).

\begin{table}[t]
    \centering
    \caption{Operator set $\mathcal{O}$ and the guidance term that measures each edit
    (\cref{sec:stage2}).}
    \label{tab:operators}
    \small
    \setlength{\tabcolsep}{4pt}
    \begin{tabularx}{\linewidth}{@{}l>{\raggedright\arraybackslash}Xl@{}}
        \toprule
        Operator & Effect on the state graph & Measured by \\
        \midrule
        \opname{Update}$(o, \alpha, s)$ & set attribute $\alpha$ (other than extent) to $s$
                                        & appearance, \cref{eq:app} \\
        \opname{Update}$(o, \alpha, s)$ & set the extent of $o$ to $s$
                                        & area, \cref{eq:ext} \\
        \opname{Link}/\opname{Unlink}$(a, r, b)$ & add/remove a planar relation
                                        & location, \cref{eq:pos} \\
        \opname{Link}/\opname{Unlink}$(a, r, b)$ & add/remove a depth relation
                                        & depth order, \cref{eq:depth} \\
        \opname{Spawn}$(o' {\leftarrow} o)$, \opname{Consume}$(o)$
                                        & add $o'$ with lineage $o$; remove $o$
                                        & presence, \cref{eq:count} \\
        \bottomrule
    \end{tabularx}
    \vspace{-5mm}
\end{table}

\parag{Event Timing Estimation.}
Physical changes are not uniform in time: an object may soften slowly and then collapse
quickly. Since VLMs are unreliable at absolute durations \citep{quantiphy2025}, the VLM only
divides the video among the events through the fractions $d_i$, a relative judgment; a slow phenomenon may thus be compressed, as in a time-lapse. We require
$\sum_{i \le K} d_i < 1$, so that the final state stays visible at the end,
and rescale the fractions otherwise. The anchor of $\mathcal{G}_i$ is the end of event~$i$; here,
$\lfloor\cdot\rceil$ denotes rounding to the nearest frame. The resulting anchors complete
the event chain in \cref{eq:interface}:
\begin{equation}
    f_i = \Bigl\lfloor \Bigl(\textstyle\sum_{j \le i} d_j\Bigr)(F-1) \Bigr\rceil,
    \qquad f_0 = 0.
    \label{eq:anchor}
\end{equation}

\subsection{Graph-Guided Test-Time Optimization}
\label{sec:stage2}

A VDM can be steered toward target frames at inference time by comparing its predicted frames
with them \citep{jang2025frame}, so a keyframe of each state is a natural target. A keyframe,
however, contains more than the graph specifies, such as the exact contour of a puddle, and
matching it pixel by pixel would impose these details on the video. We therefore compare the
video with each keyframe only on the properties selected by the edits, and update it only in
the regions of the edited objects.

\parag{Keyframe Rendering.}
Before denoising, we collect for each event the net edits $\bar{\epsilon}_i$: all edits up to
event $i$, keeping the last value of each attribute and removing edits that cancel out (\eg,
a node spawned and later consumed), which describe how $\mathcal{G}_i$ differs from
$\mathcal{G}_0$. They are written as an editing instruction, in which lineage tells where new
objects appear (\eg, a puddle where the ice was), and an image editor $\Psi$ renders the
keyframe $I_i$ from $I_0$. Editing every keyframe from $I_0$ rather than from the previous
one prevents errors from accumulating. Object masks $M_o(\cdot)$ and depth maps are then
extracted with an open-vocabulary segmenter and a monocular depth estimator.

\parag{Preliminary.}
Following Frame Guidance \citep{jang2025frame}, at denoising step $t$ the clean latent is
estimated from $z_t$; for a $v$-prediction VDM $v_\theta$, Tweedie's formula
\citep{efron2011tweedie} gives
\begin{equation}
    z_{0|t} = \sqrt{\bar\alpha_t}\, z_t - \sqrt{1-\bar\alpha_t}\, v_\theta(z_t, t),
    \label{eq:tweedie}
\end{equation}
and the corresponding estimate for flow-matching models \citep{lipman2022flow} is given in the
appendix. To save computation, only the slice of $z_{0|t}$ around each anchor is decoded into
a preview $\hat{x}_i = \mathcal{D}\bigl(z^{(f_i)}_{0|t}\bigr)$, where $z^{(f_i)}$ denotes the latent frames that cover frame $f_i$ under the temporal
compression of the VAE. The preview and the keyframe $I_i$ are then compared object by
object.

\parag{Graph-Selected Measurement.}
Comparing an object requires locating it in the preview in a form that gradients can pass
through. A segmenter cannot do this: its masks are not differentiable, and it is unreliable
on the blurry previews of early steps. We therefore locate an object $o$ by a soft occupancy
map $\Phi_o(x) \in [0,1]^{H \times W}$, which is high wherever image $x$ looks like the
object:
\begin{align}
    \Phi_o(x)(u) &= \sigma\Bigl(\bigl(\cos(\mathcal{F}(x)(u), \rho_o) - b\bigr)/\tau\Bigr),
    \label{eq:occupancy} \\
    \rho_o &= \operatorname{avg}_{M_o(I_i)} \mathcal{F}(I_i),
    \label{eq:reference}
\end{align}
where $u$ is a grid location, $\mathcal{F}$ a frozen feature encoder, $b$ and $\tau$ a
similarity threshold and a temperature, and $\operatorname{avg}_A g = \frac{1}{|A|}
\sum_{u \in A} g(u)$ the mean of a map $g$ over a region $A$. The reference feature $\rho_o$
is the mean feature of the object in the keyframe, or in $I_0$ for a consumed object.
$\Phi_o$ acts as a soft mask that changes smoothly with the preview: raising it at a location
makes the object appear there, and lowering it makes the object disappear, so the losses
below can move, grow, add, or remove an object.

Since the VDM may duplicate, lose, or split an object, we compare instances rather than whole
objects. The preview instances $\{P_j\}_{j=1}^{J}$ are the connected components of
$\{u : \Phi_o(\hat{x}_i)(u) > 0.5\}$, and the keyframe instances $\{Q_l\}_{l=1}^{L}$ those of
the mask. The Hungarian algorithm pairs them one-to-one into the matching $\pi^{*}$ of $\min(J, L)$ pairs with the lowest total centroid distance
$\sum_{(j,l) \in \pi} \lVert \bar{c}(P_j) - \bar{c}(Q_l) \rVert$, where $\bar{c}(\cdot)$ is the
centroid of an instance, and $j \notin \pi^{*}$ marks an unpaired preview instance $j$ (likewise for $l$). An unmatched instance signals a wrong number of copies, which \opname{Spawn} and \opname{Consume} change, so we measure it by suppressing unmatched preview instances and filling in unmatched keyframe ones:
\begin{equation}
\begin{aligned}
    \mathcal{L}_{\mathrm{count}}(o) = {} & \textstyle\sum_{j \notin \pi^{*}}
        \operatorname{avg}_{P_j} \Phi_o(\hat{x}_i) \\
    & + \textstyle\sum_{l \notin \pi^{*}} \max\bigl(0,\, 1 -
        \operatorname{avg}_{Q_l}\Phi_o(\hat{x}_i) / \operatorname{avg}_{Q_l}\Phi_o(I_i)\bigr).
\end{aligned}
\label{eq:count}
\end{equation}
For example, a consumed object has no keyframe instance, so all its preview instances are
suppressed, and a spawned object missing from early previews is filled in.

Each matched pair is then compared only on the property that its edit changes. With
$R_{jl} = P_j \cup Q_l$, which covers where the object is and where it should be, the terms
are, summing over $(j,l) \in \pi^{*}$,
\begin{subequations}
\label{eq:terms}
\begin{align}
    \mathcal{L}_{\mathrm{app}} &= \textstyle\sum 1 - \cos\bigl(
        \bar{\mathcal{F}}_{P_j}(\hat{x}_i),\, \bar{\mathcal{F}}_{Q_l}(I_i) \bigr),
        \label{eq:app} \\
    \mathcal{L}_{\mathrm{ext}} &= \textstyle\sum \log^2 \bigl( \lvert \Phi_o(\hat{x}_i)
        {\odot} R_{jl} \rvert / \lvert \Phi_o(I_i) {\odot} R_{jl} \rvert \bigr),
        \label{eq:ext} \\
    \mathcal{L}_{\mathrm{pos}} &= \textstyle\sum \bigl\lVert
        c_{jl}(\hat{x}_i) - c_{jl}(I_i) \bigr\rVert^2,
        \label{eq:pos} \\
    \mathcal{L}_{\mathrm{depth}} &= \bigl( \Delta z_{ab}(\hat{x}_i) - \Delta z_{ab}(I_i)
        \bigr)^2,
        \label{eq:depth}
\end{align}
\end{subequations}
where $\bar{\mathcal{F}}_A(x) = \operatorname{avg}_A \mathcal{F}(x)$, $\lvert\cdot\rvert$ sums
a map, $c_{jl}(x)$ is the centroid of $\Phi_o(x)$ within $R_{jl}$, and $\Delta z_{ab}$ is the
difference between the mean normalized depths of $a$ and $b$, weighted by occupancy in the
preview. The terms measure appearance, area as a scale-invariant ratio, location, and depth
order. Area and location are compared on $R_{jl}$ rather than on the preview mask, since a
loss only changes the occupancy inside its region and the object must be able to grow or
move toward its target. For an edit on a pair $(a, b)$, the term is computed for both
objects. Since every term compares an object-level summary, not pixels, it ignores the exact
contour drawn by the editor.

The objective at event $i$ sums the term selected by each edit and a count term for every
object:
\begin{equation}
    \mathcal{L}^{(i)} = \sum_{\epsilon \in \bar{\epsilon}_i}
    \mathcal{L}_{\mu(\epsilon)}(\epsilon)
    + \sum_{o \in \mathcal{V}_{\le i}} \mathcal{L}_{\mathrm{count}}(o),
    \label{eq:objective}
\end{equation}
where $\mu(\epsilon) \in \{\mathrm{app}, \mathrm{ext}, \mathrm{pos}, \mathrm{depth}\}$ is the
term selected by edit $\epsilon$ (\cref{tab:operators}), and $\mathcal{V}_{\le i}$ contains
every object that has appeared up to event $i$. The first sum excludes \opname{Spawn} and
\opname{Consume}, which the second measures; the second also covers unchanged objects, so the
VDM neither loses nor duplicates them. We write $\mathcal{L}^{(i)}_k$ for the $k$-th term;
all terms have equal weight, since each gradient is normalized (\cref{eq:update}). Properties
that the graph does not select give no gradient: a melting event is judged on the material
and size of the object, not on the shape of the liquid.

\parag{Region-Weighted Update.}
Although each term looks only at its objects, its gradient spreads over the whole latent
through the large receptive fields of the decoder and denoiser, and would change regions that should stay fixed. We therefore restrict each term to a region built from its instance masks,
\begin{equation}
    \Omega_k = \bigcup_{(j,l) \in \pi^{*}} \mathrm{hull}\bigl(P_j \cup Q_l\bigr)
    \;\cup \bigcup_{j \notin \pi^{*}} P_j
    \;\cup \bigcup_{l \notin \pi^{*}} Q_l,
    \label{eq:region}
\end{equation}
where $\mathrm{hull}(\cdot)$ is the convex hull, covering where a matched object is, where it
should be, and the space between, and the unmatched instances are included so that they can
be suppressed or filled in. $\Omega_k$ is average-pooled to the latent grid, and the update is
\begin{equation}
    z_t \leftarrow z_t - \eta \textstyle\sum_{i=1}^{K} \sum_{k}
    \bigl( \Omega_k + \lambda (1 - \Omega_k) \bigr) \odot
    \frac{\nabla_{z_t} \mathcal{L}^{(i)}_k}{\lVert \nabla_{z_t} \mathcal{L}^{(i)}_k \rVert},
    \label{eq:update}
\end{equation}
where $\eta$ is the step size and $\lambda \in [0,1]$ reduces the update outside the region
($\lambda = 1$ gives unweighted guidance). As in Frame Guidance, gradients are normalized to
unit norm, here per term, since the terms have different scales. The same spatial region is
applied at every temporal index: the gradient at an anchor reaches all frames through the
denoiser, and the weighting limits it in space but not in time. Following Frame Guidance,
\cref{eq:update} is applied $N_{\mathrm{g}}$ times per step while the layout forms
($t > t_E$), the unweighted gradient is applied with re-noising for $t_E \ge t > t_L$, and the
remaining steps are unguided. Details are in \cref{app:guidance}.

%% file: sections/experiment_v2.tex
\section{Experiments}
\label{sec:experiments}

\subsection{Setup}
\label{subsec:implementation_details}

\parag{Implementation.}
GPSR uses Gemini 3 Flash \citep{team2023gemini} with JSON-constrained outputs. Keyframes are
rendered with Gemini 3 Pro Image \citep{team2023gemini}, masks with Grounded-SAM-2
\citep{ravi2025sam}, depth with Depth Anything V2 \citep{yang2024depth}, and occupancy with
DINOv3 features \citep{simeoni2025dinov3}. The frozen VDM is CogVideoX-I2V-5B
\citep{yang2025cogvideox}, with guidance defaults from Frame Guidance \citep{jang2025frame} (\cref{app:hyperparameters}).

\parag{Benchmarks and baselines.}
PhyGenBench \citep{meng2024towards} covers 27 phenomena in four domains; since it is a
text-to-video (T2V) benchmark, we generate an initial frame for each prompt with the Gemini
API \citep{team2023gemini}, shared by all I2V methods. Physics-IQ \citep{motamed2025generative}
contains 396 real-world scenarios in five domains. We compare with I2V models
(CogVideoX-I2V-5B, SVD-XT \citep{blattmann2023stable}, LTX-Video-I2V \citep{hacohen2024ltx}),
VLIPP \citep{yang2025vlipp} (reported), and Frame Guidance \citep{jang2025frame}, which we run
with the same model, protocol, anchors, and keyframes as PhysPlan but with its original L2
loss to the whole keyframe. On PhyGenBench we also list T2V foundation models
\citep{zheng2024open,yang2025cogvideox,wan2025wan,team2025kling} and physics-aware methods
\citep{xue2025phyt2v,chen2026hierarchical,wang2026physically,cai2026phygdpo,causalmotion2026}
as reported in prior work (full list in \cref{app:full_results}).

\begin{table}[!t]
    \centering
    \caption{\textbf{Physical plausibility} on PhyGenBench (PCA score in $[0, 1]$) and
    Physics-IQ; higher is better, and averages are weighted by the number of samples per
    domain. Best in \textbf{bold}, second best \underline{underlined}; subscripts give the
    gain of PhysPlan over its base model, CogVideoX-I2V-5B. T2V results are reported by
    \citet{wang2026physically}, except LTX-Video and CausalMotion \citep{causalmotion2026};
    VLIPP is reported by \citet{yang2025vlipp}.}
    \label{tab:main}
    \setlength{\tabcolsep}{4pt}
    \resizebox{\textwidth}{!}{%
    \begin{tabular}{l ccccl c cccccl}
        \toprule
        & \multicolumn{5}{c}{\textbf{PhyGenBench}} && \multicolumn{6}{c}{\textbf{Physics-IQ}} \\
        \cmidrule(lr){2-6} \cmidrule(lr){8-13}
        \textbf{Model} & Mech. & Optics & Thermal & Material & \textbf{Avg.}
            && S.M. & F.D. & Optics & Magn. & Thermo. & \textbf{Avg.} \\
        \midrule

        \multicolumn{13}{l}{\textit{Text-to-video models}} \\
        Kling
            & 0.45 & 0.58 & 0.50 & 0.40 & 0.49
            & \multicolumn{6}{c}{} \\
        Wan2.2-14B
            & 0.53 & 0.61 & 0.58 & 0.43 & 0.54
            & \multicolumn{6}{c}{} \\
        CogVideoX-5B
            & 0.39 & 0.55 & 0.40 & 0.42 & 0.45 & 
            & \multicolumn{6}{c}{\textit{Physics-IQ is evaluated for I2V models only}} \\
        \quad + PhysHPO
            & 0.55 & 0.68 & 0.50 & \second{0.65} & 0.61
            & \multicolumn{6}{c}{} \\
        \quad + Chain-of-Events
            & \second{0.70} & \best{0.79} & \best{0.77} & 0.64 & \second{0.73}
            & \multicolumn{6}{c}{} \\
        LTX-Video
            & 0.35 & 0.45 & 0.36 & 0.38 & 0.39
            & \multicolumn{6}{c}{} \\
        \quad + CausalMotion
            & 0.61 & 0.71 & 0.68 & 0.61 & 0.65
            & \multicolumn{6}{c}{} \\
        \midrule
        \multicolumn{13}{l}{\textit{Image-to-video models}} \\
        CogVideoX-I2V-5B
            & 0.48 & 0.69 & 0.43 & 0.41 & 0.52
            && 30.4 & 29.8 & 16.7 & 13.3 & 8.5 & 27.1 \\
        SVD-XT
            & 0.46 & 0.68 & 0.48 & 0.41 & 0.52
            && 21.9 & 20.5 & 6.8 & 8.4 & \best{17.1} & 19.1 \\
        LTX-Video-I2V
            & 0.47 & 0.65 & 0.46 & 0.37 & 0.50
            && 30.2 & 29.8 & 15.9 & 13.2 & 8.4 & 26.8 \\

        \midrule
        \multicolumn{13}{l}{\textit{Reasoning-guided and training-free guided I2V methods}} \\
        VLIPP
            & 0.55 & 0.71 & 0.60 & 0.53 & 0.60
            && \second{42.3} & \best{34.1} & 16.9 & 13.4 & 8.8 & \second{34.6} \\
        Frame Guidance
            & 0.52 & 0.56 & 0.47 & 0.48 & 0.51
            && 35.4 & 27.4 & \second{24.1} & \second{13.9} & 8.4 & 30.3 \\
        \rowcolor{oursrow}
        \textbf{PhysPlan (Ours)}
            & \best{0.81} & \second{0.78} & \second{0.74} & \best{0.75}
            & \best{0.77}\gain{+0.25}
            && \best{45.6} & \second{31.8} & \best{30.4} & \best{19.7}
            & \second{9.5} & \best{38.2}\gain{+11.1} \\

        \bottomrule
    \end{tabular}%
    }
    \vspace{-3mm}
\end{table}

\subsection{Physical Plausibility}
\label{subsec:quantitative}

\parag{PhyGenBench.}
PhysPlan reaches 0.77 on average (\cref{tab:main}), against 0.52 for its base model and 0.60
for VLIPP, the strongest prior I2V method. The largest gains are in Material ($+0.34$),
Mechanics ($+0.33$), and Thermal ($+0.31$), which are dominated by changes of object state
that the graph expresses directly as edits. The gain in Optics is smaller ($+0.09$), since
optical phenomena change how light passes through the scene rather than object states. Frame
Guidance, which matches the same keyframes as whole frames, does not improve over its base
model (0.51 vs.\ 0.52), whereas PhysPlan gains $+0.26$ from them. The T2V event chain of
\citet{wang2026physically} reaches 0.73 and is ahead in Optics and Thermal, though T2V and I2V
results are not directly comparable.

\parag{Physics-IQ.}
PhysPlan reaches 38.2, against 27.1 for its base model, 30.3 for Frame Guidance, and 34.6 for
VLIPP. The largest gains are in Solid Mechanics ($+15.2$), Optics ($+13.7$), and Magnetism
($+6.4$), where the outcome depends on which objects move or change while the rest of these
cluttered real scenes stays fixed. Frame Guidance improves over its base model but lowers
Fluid Dynamics and Thermodynamics, whereas PhysPlan improves in every domain. It remains
below VLIPP in Fluid Dynamics (31.8 vs.\ 34.1), since continuous flow between anchors is left
to the VDM while VLIPP plans trajectories, and all guided methods stay near the base model in Thermodynamics (9 videos).

\parag{Qualitative Evaluation.}
\label{subsec:qualitative}

\begin{figure*}[t!]
    \centering
    \includegraphics[width=\textwidth]{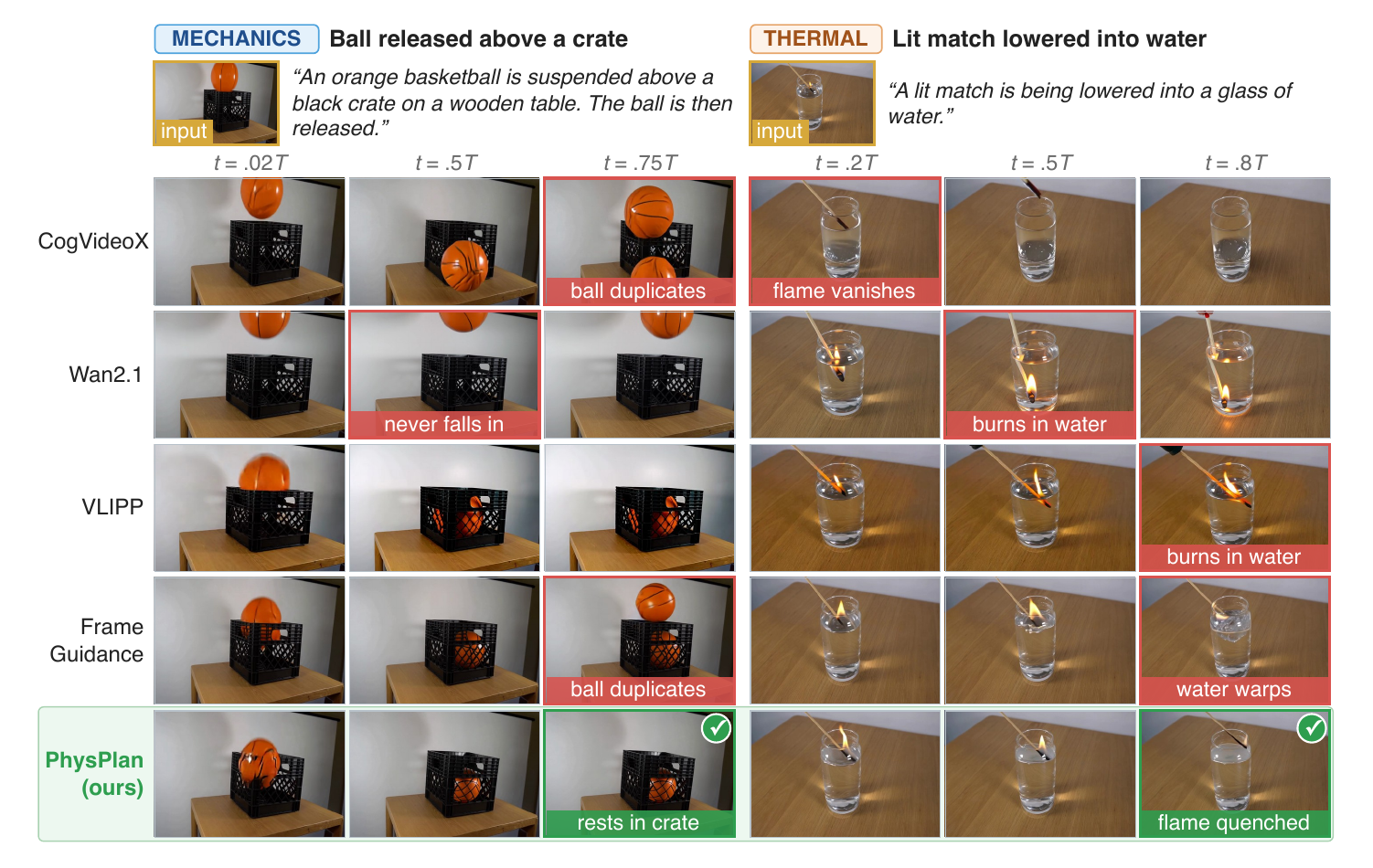}
    \vspace{-4mm}
    \caption{\textbf{Qualitative comparison.} Frames at matched times $t/T$ for a mechanics
    (\emph{left}) and a thermal (\emph{right}) scene. Red marks a physical error, green the
    correct outcome: only PhysPlan keeps a single ball that lands in the crate, and only
    PhysPlan extinguishes the flame when the match enters the water.}
    \label{fig:qualitative}
    \vspace{-3mm}
\end{figure*}

In \cref{fig:qualitative}, the baselines duplicate the ball, let it miss the crate, keep the
match burning under water, or warp the water. PhysPlan follows its event chain in both scenes:
the single ball comes to rest in the crate, and the flame is quenched as the match enters the
water.

\subsection{Visual Quality}
\label{subsec:visual_quality}

\begin{wraptable}{r}{0.5\linewidth}
    \vspace{-5mm}
    \centering
    \caption{Perceptual quality on PhyGenBench and Physics-IQ, measured by FID and FVD.
    Best results are in \textbf{bold}.}
    \label{tab:visual}
    \resizebox{0.5\columnwidth}{!}{%
    \begin{tabular}{l c c c c}
        \toprule
        \multirow{2}{*}{\textbf{Model}}
        & \multicolumn{2}{c}{\textbf{PhyGenBench}}
        & \multicolumn{2}{c}{\textbf{Physics-IQ}} \\
        \cmidrule(lr){2-3} \cmidrule(lr){4-5}
        & FID$\downarrow$ & FVD$\downarrow$
        & FID$\downarrow$ & FVD$\downarrow$ \\
        \midrule
        CogVideoX-I2V-5B & 48.2 & 632.8 & 55.4 & 698.5 \\
        Frame Guidance & 46.4 & 580.4 & 49.2 & 603.4 \\
        \rowcolor{oursrow}
        \textbf{PhysPlan (Ours)} & \textbf{45.4} & \textbf{500.2} & \textbf{47.7} & \textbf{495.6} \\
        \bottomrule
    \end{tabular}%
    }
    \vspace{-2mm}
\end{wraptable}

Guidance could trade visual quality for plausibility, so we evaluate it in three ways. On
\emph{distribution metrics} (\cref{tab:visual}), PhysPlan improves FID \citep{heusel2017gans}
over its base model and Frame Guidance, and reduces FVD \citep{ge2024content} by
about 21\% and 29\%, which we attribute to the region-weighted update keeping the background
stable. On \emph{per-video scores} from VBench \citep{huang2024vbench} (\cref{tab:vbench} in the appendix),
which do not penalize deviations from a reference distribution, PhysPlan raises the Quality
Score from 83.05 to 84.88 and improves six of seven dimensions, most in Dynamic Degree
($+6.11$); it also exceeds closed-source VDMs in Quality Score and Dynamic Degree. In a two-alternative forced-choice \emph{user study} with 60 participants against CogVideoX-I2V-5B or Frame Guidance, PhysPlan was preferred in 72\%, 60\%, and 73\% of the
comparisons for physical plausibility, frame quality, and temporal smoothness
(\cref{app:user_study}). Plausibility thus does not cost visual quality.

\subsection{Ablation Study}
\label{subsec:ablation}

\textbf{(i) Prompt-derived plan}: no state graph; the VLM describes each state as a complete
scene, used as the keyframe instruction, with the whole-frame L2 loss. \textbf{(ii)
Full-graph regeneration}: the VLM rewrites the full graph from each delta, and the edits are
taken as differences between graphs. \textbf{(iii) Whole-frame L2 loss}: the graph-selected
terms are replaced by the L2 loss of Frame Guidance without region weighting (the Frame Guidance baseline). \textbf{(iv) Without region weighting}: $\lambda = 1$ in
\cref{eq:update}. \textbf{(v)--(viii)}: one term of \cref{eq:terms} is removed, leaving its edits (\cref{tab:operators}) unmeasured.

\begin{table}[!t]
    \centering
    \caption{Ablation study on Physics-IQ. Each setting changes one component of PhysPlan
    and keeps all others fixed. S.M.\ refers to Solid Mechanics
    and F.D.\ to Fluid Dynamics; the average is weighted by the number of videos per
    category, and subscripts give its change from the full model. Best results are in \textbf{bold}.}
    \label{tab:ablation}
    \resizebox{\columnwidth}{!}{%
    \begin{tabular}{l l c c c c c l}
        \toprule
        & \textbf{Setting} & \textbf{S.M.} ($\uparrow$) & \textbf{F.D.} ($\uparrow$) &
            \textbf{Optics} ($\uparrow$) & \textbf{Magn.} ($\uparrow$) &
            \textbf{Thermo.} ($\uparrow$) & \textbf{Avg.} ($\uparrow$) \\
        \midrule
        \rowcolor{oursrow}
        & \textbf{PhysPlan (full)} & \textbf{45.6} & \textbf{31.8} & \textbf{30.4} &
            \textbf{19.7} & \textbf{9.5} & \textbf{38.2} \\
        \midrule
        \multicolumn{8}{l}{\textit{Grounded Physical State Reasoning}} \\
        (i)     & Prompt-derived plan     & 31.1 & 30.2 & 20.4 & 13.8 & 8.6 & 28.1\drop{$-$10.1} \\
        (ii)    & Full-graph regeneration & 35.2 & 30.2 & 23.6 & 15.9 & 8.9 & 30.9\drop{$-$7.3} \\
        \midrule
        \multicolumn{8}{l}{\textit{Graph-Guided Test-Time Optimization}} \\
        (iii)   & Whole-frame L2 loss (= Frame Guidance) & 35.4 & 27.4 & 24.1 & 13.9 & 8.4 & 30.3\drop{$-$7.9} \\
        (iv)    & w/o region weighting ($\lambda = 1$) & 42.8 & 30.7 & 28.3 & 17.1 & 9.1 & 36.0\drop{$-$2.2} \\
        \midrule
        \multicolumn{8}{l}{\textit{Measurement terms}} \\
        (v)     & w/o $\mathcal{L}_{\mathrm{app}}$   & 42.6 & 29.4 & 25.5 & 18.1 & 9.3 & 35.3\drop{$-$2.9} \\
        (vi)    & w/o $\mathcal{L}_{\mathrm{ext}}$   & 43.1 & 30.6 & 28.7 & 18.8 & 8.8 & 36.2\drop{$-$2.0} \\
        (vii)   & w/o $\mathcal{L}_{\mathrm{pos}}$   & 40.7 & 31.2 & 29.2 & 17.5 & 9.2 & 35.0\drop{$-$3.2} \\
        (viii)  & w/o $\mathcal{L}_{\mathrm{depth}}$ & 42.9 & 31.2 & 24.2 & 15.4 & 8.9 & 35.6\drop{$-$2.6} \\
        \bottomrule
    \end{tabular}%
    }
\end{table}

\parag{Results.}
Every component contributes, and the two gaps of \cref{sec:intro} cause the largest drops
(\cref{tab:ablation}). Removing the state graph (i) costs 10.1 points, of which grounding alone
accounts for 2.2 when compared with (iii), which uses the same loss. Regenerating the graph
(ii) loses most in Solid Mechanics, since unmentioned objects can then change; matching the
same keyframes as whole frames (iii) loses most in Solid Mechanics and Optics, where
incidental keyframe details are imposed on the video; and removing the region weighting (iv)
costs 2.2 points. Each term matters most where it is designed to: $\mathcal{L}_{\mathrm{pos}}$
for Solid Mechanics, where outcomes depend on where objects move and come to rest;
$\mathcal{L}_{\mathrm{app}}$ for Fluid Dynamics and Optics, whose outcomes are visible as
changes of appearance such as wet or transparent surfaces; and $\mathcal{L}_{\mathrm{depth}}$
for Optics and Magnetism, whose outcomes often depend on which object is in front of or
inside another.

%% file: sections/6.conclusion.tex
\section{Conclusion}
\label{sec:conclusion}

We presented PhysPlan, a training-free framework that turns physical reasoning about the
observed scene into guidance for a frozen VDM. GPSR grounds the phenomenon in a state graph
of the observed frame and evolves it only through verified edits, so every change is
traceable and every other object stays unchanged. GTO uses the resulting event chain to
decide what to measure, where to update, and when each state should hold, comparing the
video with keyframes object by object and only along the properties the edits select.
PhysPlan achieves the best average scores on PhyGenBench (0.77) and Physics-IQ (38.2), far
above its base model (0.52 and 27.1), and with the same keyframes it outperforms Frame
Guidance by 0.26 and 7.9 points, while also improving visual quality and being preferred by
participants in a user study. The ablations confirm both gaps that motivate its design. Our
results suggest that the physical knowledge of VLMs is most useful to video generation when
expressed as explicit, verifiable changes to the observed scene, which tell the generator
both what should happen and what it should leave alone.

\parag{Limitations and future work.}
\label{sec:limitations}
PhysPlan represents a phenomenon through discrete objects: every node, edit, and
measurement refers to an object with an instance, a mask, and a centroid, and the video is
constrained only at the anchor frames. This makes it most effective when a phenomenon is
carried by clearly delineated objects, where it achieves its largest gains, \eg, in Solid
Mechanics, Optics, and Magnetism on Physics-IQ and in Mechanics and Material on PhyGenBench.
It is weaker in Fluid Dynamics, where liquids that spread, splash, or split lack clear object
boundaries and their continuous motion between anchors is left to the VDM. VLIPP stays ahead
there by planning trajectories, although its continuous motion is interpolated from a few
coarse waypoints and may itself deviate from the true dynamics. Future work includes
representations for phenomena beyond discrete objects, such as fields or particle sets for
fluids, combined with motion planning between anchors, as well as evaluation on further VDMs.

%% file: sections/7.appendix.tex
\providecommand{\opname}[1]{\textsc{#1}}

\begin{center}
    {\Large\bfseries Appendix}
\end{center}
\vspace{2mm}
\noindent This appendix provides the method details, additional quantitative results, failure
and cost analyses, and the user study for PhysPlan.

\vspace{4mm}
{\setlength{\parskip}{0pt}
\appsec{app:method}
    \appsub{app:vocab}
    \appsub{app:checks}
    \appsub{app:prompts}
    \appsub{app:keyframe}
    \appsub{app:guidance}
    \appsub{app:loss}
    \appsub{app:example}
\appsec{app:full_results}
\appsec{app:failure}
\appsec{app:cost}
\appsec{app:user_study}
}
\clearpage

\section{Method Details}
\label{app:method}

This section gives the details of Grounded Physical State Reasoning (\cref{sec:stage1}) and
Graph-Guided Test-Time Optimization (\cref{sec:stage2}) that are omitted from the main paper. The
notation follows the main paper.

\subsection{State Graph Vocabulary}
\label{app:vocab}

\Cref{tab:app_attributes} defines the six attribute keys of a node. Keys are fixed and
values are open-vocabulary; the examples are illustrative, not exhaustive. The boundary
between attributes matters for the checks, so we fix it as follows. Integrity describes
damage that keeps the object as one object; once the object separates into parts, the parts
are spawned as new objects. Surface describes the outer layer without a change of the
object itself: a wet tray is a surface change, whereas softening butter is a change of
material phase. Rust, charring, and dirt are recorded as surface changes, and color is
reserved for changes of hue alone.

\begin{table}[!h]
    \centering
    \caption{\textbf{Attribute keys of a node.} All attributes except extent are measured
    by the appearance term $\mathcal{L}_{\mathrm{app}}$; extent is measured by the area term
    $\mathcal{L}_{\mathrm{ext}}$.}
    \label{tab:app_attributes}
    \small
    \begin{tabularx}{\linewidth}{@{}l>{\raggedright\arraybackslash}X>{\raggedright\arraybackslash}X@{}}
        \toprule
        \rowcolor{gray!20} \textbf{Key} & \textbf{Describes} & \textbf{Example values} \\
        \midrule
        material phase & physical phase of the material & solid, softened, partially
            melted, liquid, gas \\
        integrity      & structural wholeness            & intact, cracked, dented, torn,
            crushed \\
        surface        & condition of the outer layer    & dry, wet, frosted, fogged,
            rusted, charred \\
        color          & hue of the object               & green, brown, red \\
        extent         & size of the object              & small, large, spread over most
            of the tray \\
        configuration  & pose or shape                   & upright, tipped over, open,
            closed, bent, crumpled \\
        \bottomrule
    \end{tabularx}
\end{table}

\Cref{tab:app_relations} defines the relation types. Spatial relations are written in one
direction only (left of, above, in front of), so that the same layout is never recorded in
two ways. The graph records states, not processes: actions such as pouring, heating, or
pushing are not relations, and appear only through their results, such as a spawned stream
of water, a changed phase, or a new support relation. Common situations that do not match
a relation name are mapped to the fixed set by the conventions in
\cref{tab:app_conventions}, which are included in the parsing instruction.

\begin{table}[!h]
    \centering
    \caption{\textbf{Relation types.} Planar relations are measured by the location term
    $\mathcal{L}_{\mathrm{pos}}$, depth relations by the depth term
    $\mathcal{L}_{\mathrm{depth}}$.}
    \label{tab:app_relations}
    \small
    \begin{tabularx}{\linewidth}{@{}lll>{\raggedright\arraybackslash}X@{}}
        \toprule
        \rowcolor{gray!20} \textbf{Relation $(a, r, b)$} & \textbf{Kind} &
            \textbf{Measured as} & \textbf{Meaning} \\
        \midrule
        support     & physical & planar & $a$ holds $b$ up against gravity \\
        contact     & physical & planar & $a$ and $b$ touch \\
        containment & physical & depth  & $b$ is inside $a$ \\
        attachment  & physical & planar & $b$ is fixed to $a$ \\
        left of     & spatial  & planar & $a$ is left of $b$ in the image \\
        above       & spatial  & planar & $a$ is above $b$ in the image \\
        in front of & spatial  & depth  & $a$ is closer to the camera than $b$ \\
        near        & spatial  & planar & $a$ is close to $b$ without contact \\
        \bottomrule
    \end{tabularx}
\end{table}

\begin{table}[!h]
    \centering
    \caption{\textbf{Conventions for mapping common situations to relations.}}
    \label{tab:app_conventions}
    \small
    \begin{tabularx}{\linewidth}{@{}>{\raggedright\arraybackslash}X>{\raggedright\arraybackslash}X@{}}
        \toprule
        \rowcolor{gray!20} \textbf{Situation} & \textbf{Relations} \\
        \midrule
        object floating on a liquid (ice in water, a boat on a lake) & support by the liquid \\
        object submerged in a liquid (a stone in water, a teabag in a cup) & containment \\
        object leaning against another (a ladder against a wall) & support + contact \\
        object hanging from another (a lamp from a ceiling) & attachment \\
        part of an object (a handle on a cup) & attachment \\
        layer covering an object (snow on a car, a cloth over a table) & support + surface
            change of the covered object \\
        right of, below, behind & inverse of left of, above, in front of \\
        \bottomrule
    \end{tabularx}
\end{table}

\subsection{Deterministic Checks}
\label{app:checks}

Let $\mathcal{N}(\mathcal{S}_i)$ be the objects listed in $\mathcal{S}_i$ and
$\mathcal{N}^{+}(\epsilon_i)$ the nodes spawned by $\epsilon_i$. An edit set $\epsilon_i$
is accepted if it passes all four checks.
\begin{enumerate}[leftmargin=*]
    \item \textbf{Grounding.} Every node referenced by an edit is in $\mathcal{V}_{i-1}$ or
    in $\mathcal{N}^{+}(\epsilon_i)$; every attribute key is one of the six keys of
    \cref{tab:app_attributes}; every relation type is one of \cref{tab:app_relations};
    every spawned identifier is new.
    \item \textbf{Coverage.} Every object in $\mathcal{N}(\mathcal{S}_i)$ is the target of
    at least one edit. \opname{Update} and \opname{Consume} target only objects in
    $\mathcal{N}(\mathcal{S}_i)$, and every \opname{Link} or \opname{Unlink} has at least
    one endpoint in $\mathcal{N}(\mathcal{S}_i) \cup \mathcal{N}^{+}(\epsilon_i)$. The
    other endpoint may be an unlisted object, since adding a relation (\eg, the tray now
    supports the puddle) does not change that object's attributes.
    \item \textbf{Lineage.} Every \opname{Spawn} names a source node in $\mathcal{V}_{i-1}$;
    nodes are added only by \opname{Spawn} and removed only by \opname{Consume}; no edit
    acts on a node consumed earlier in the chain.
    \item \textbf{Consistency.} No attribute of a node is set twice in $\epsilon_i$; no
    edge is both linked and unlinked; after applying $\epsilon_i$, the support and
    containment relations form directed acyclic graphs.
\end{enumerate}
If a check fails, the list of violations is returned to the VLM with $p_{\mathrm{edit}}$,
for at most 3 retries. If the edits are still rejected, the delta is regenerated.

\subsection{VLM Prompts}
\label{app:prompts}

GPSR uses a single VLM (Gemini 3 Flash) with three instructions: scene parsing
($p_{\mathrm{parse}}$), phenomenon decomposition ($p_{\mathrm{delta}}$), and delta
translation ($p_{\mathrm{edit}}$). The \textcolor{promptvar}{\textbf{colored texts}}
indicate dynamic inputs. All outputs are constrained to strict JSON schemas, which we parse
and validate before use.

\addvspace{1em}

\raggedbottom
\noindent\textbf{Prompt 1: Scene Parsing} ($p_{\mathrm{parse}}$):\\
This instruction produces the state graph $\mathcal{G}_0$.

\begin{promptbox}
You are an expert in physical scene understanding. Your task is to describe the objects in
an image and their physical relations as a state graph.

\textbf{Inputs:}
\begin{itemize}[leftmargin=*, nosep]
    \item Initial Image: \textcolor{promptvar}{\textbf{<INITIAL\_FRAME>}}
    \item User Prompt: \textcolor{promptvar}{\textbf{<TEXT\_PROMPT>}}
\end{itemize}

\textbf{Step 1: Objects.}
List every object visible in the image, including objects that the prompt does not
mention. Give each object an identifier of the form \texttt{<category>\#<n>}, a category,
and a value for each of the attribute keys: material phase, integrity, surface, color,
extent, configuration.

\textbf{Step 2: Relations.}
List all relations $(a, r, b)$ between objects, with $r$ from: support, contact,
containment, attachment, left of, above, in front of, near. Follow these conventions:
\textcolor{promptvar}{\textbf{<RELATION\_CONVENTIONS>}}.

\textbf{Output Format \& Example:} \\
Return a structured JSON object exactly matching this schema:
\begin{lstlisting}[language=json, backgroundcolor=\color{gray!10}, xleftmargin=1em]
{
  "nodes": [
    {"id": "ice#1", "category": "ice cube",
     "attributes": {"material_phase": "solid", "integrity": "intact",
                    "surface": "dry", "color": "clear",
                    "extent": "small", "configuration": "upright"}},
    {"id": "tray#2", "category": "tray", "attributes": {...}}
  ],
  "edges": [
    {"a": "tray#2", "r": "support", "b": "ice#1"}
  ]
}
\end{lstlisting}
\end{promptbox}

\addvspace{1em}
\noindent\textbf{Prompt 2: Phenomenon Decomposition} ($p_{\mathrm{delta}}$):\\
This instruction produces the physical deltas $\Delta = (\delta_1, \dots, \delta_K)$.

\begin{promptbox}
You are an expert in physical reasoning. Your task is to predict how a scene changes during
a physical phenomenon, as a sequence of states.

\textbf{Inputs:}
\begin{itemize}[leftmargin=*, nosep]
    \item Initial Image: \textcolor{promptvar}{\textbf{<INITIAL\_FRAME>}}
    \item Scene Graph: \textcolor{promptvar}{\textbf{<GRAPH\_0>}}
    \item User Prompt: \textcolor{promptvar}{\textbf{<TEXT\_PROMPT>}}
\end{itemize}

\textbf{Step 1: Events.}
Break the phenomenon into a sequence of events in causal order. Objects that change at the
same time belong to the same event.

\textbf{Step 2: States and rules.}
For each event, list every object whose state changes, using the identifiers of the scene
graph. For each object, give its new state in words and the physical rule that produces
it, as a short qualitative statement without numbers. Include effects on objects that the
prompt does not mention. If an object comes into existence, give it a new identifier and
name the object it comes from. Objects you do not list stay unchanged.

\textbf{Step 3: Timing.}
Give the fraction of the video that each event takes. The fractions must sum to less than
1, so that the final state remains visible at the end of the video.

\textbf{Output Format \& Example:} \\
Return a structured JSON object exactly matching this schema:
\begin{lstlisting}[language=json, backgroundcolor=\color{gray!10}, xleftmargin=1em]
{
  "deltas": [
    {"states": [
       {"object": "ice#1", "state": "partially melted",
        "rule": "ice above its melting point turns into water"},
       {"object": "tray#2", "state": "wet beneath ice#1",
        "rule": "liquid flows down onto its support"}],
     "fraction": 0.30},
    {"states": [
       {"object": "ice#1", "state": "melted into puddle#3 on tray#2",
        "rule": "ice above its melting point turns into water",
        "new_object": {"id": "puddle#3", "source": "ice#1"}}],
     "fraction": 0.30}
  ]
}
\end{lstlisting}
\end{promptbox}

\addvspace{1em}
\noindent\textbf{Prompt 3: Delta Translation} ($p_{\mathrm{edit}}$):\\
This instruction translates one delta into typed edits $\epsilon_i$.

\begin{promptbox}
You are a precise graph editor. Your task is to express a change of state as edits to a
scene graph.

\textbf{Inputs:}
\begin{itemize}[leftmargin=*, nosep]
    \item Current Scene Graph: \textcolor{promptvar}{\textbf{<GRAPH\_i-1>}}
    \item Physical Delta: \textcolor{promptvar}{\textbf{<DELTA\_i>}}
    \item Violations of the previous attempt (on retry only):
          \textcolor{promptvar}{\textbf{<VIOLATIONS>}}
\end{itemize}

\textbf{Step 1: Edits.}
Express the delta using only these operators: \texttt{Update(o, key, value)},
\texttt{Link(a, r, b)}, \texttt{Unlink(a, r, b)},
\texttt{Spawn(new\_id <- source\_id, category, attributes)}, and \texttt{Consume(o)}.

\textbf{Step 2: Scope.}
Edit exactly the objects listed in the delta. Use only the attribute keys and relation
types of the scene graph. If violations are given, correct them.

\textbf{Output Format \& Example:} \\
Return a structured JSON object exactly matching this schema:
\begin{lstlisting}[language=json, backgroundcolor=\color{gray!10}, xleftmargin=1em]
{
  "edits": [
    {"op": "Spawn", "id": "puddle#3", "source": "ice#1",
     "category": "puddle", "attributes": {"material_phase": "liquid", ...}},
    {"op": "Consume", "o": "ice#1"},
    {"op": "Link", "a": "tray#2", "r": "support", "b": "puddle#3"}
  ]
}
\end{lstlisting}
\end{promptbox}
\flushbottom

\subsection{Keyframe Rendering}
\label{app:keyframe}

For each event $i$, the net edits $\bar{\epsilon}_i$ describe how $\mathcal{G}_i$ differs
from $\mathcal{G}_0$. They are translated into a natural-language editing instruction $c_i$
by the same VLM with the prompt $p_{\mathrm{render}}$ below. The translation replaces node
identifiers with referring expressions that the image editor can resolve, using the category
and, when a category occurs more than once, a relation from $\mathcal{G}_0$ (\eg, ``the ice
cube on the tray''); it uses lineage to place new objects (\eg, a puddle where the ice was);
and it describes only the edited objects, ending with an explicit request to keep everything
else unchanged. Since every keyframe is edited from $I_0$, the instruction describes the full
change from the observed frame to state $i$, not the change from the previous state. For the
final state of \cref{tab:app_example}, $c_3$ is: ``Replace the ice cube with a puddle of water
that spreads over most of the tray, and make the tray wet around it. Keep everything else in
the image unchanged.''

\addvspace{1em}
\noindent\textbf{Prompt 4: Keyframe Instruction} ($p_{\mathrm{render}}$):\\
This instruction translates the net edits $\bar{\epsilon}_i$ into the editing instruction
$c_i$.

\begin{promptbox}
You are writing an instruction for an image editor. The editor receives the initial image
and your instruction, and must change the image so that it shows a later state of the
scene. The editor does not know the scene graph or the identifiers of objects.

\textbf{Inputs:}
\begin{itemize}[leftmargin=*, nosep]
    \item Initial Image: \textcolor{promptvar}{\textbf{<INITIAL\_FRAME>}}
    \item Initial Scene Graph: \textcolor{promptvar}{\textbf{<GRAPH\_0>}}
    \item Net Edits: \textcolor{promptvar}{\textbf{<NET\_EDITS\_i>}}
\end{itemize}

\textbf{Step 1: Refer to objects.}
Replace every identifier with a short description that picks out the object in the image,
using its category and, if the category occurs more than once, its relation to another
object (\eg, ``the ice cube on the tray'').

\textbf{Step 2: Describe the changes.}
Write one clause per edit, describing the state the object should be in, not the process
that leads to it. For a new object, say where it appears, using the object it comes from
(\eg, ``a puddle where the ice cube was''). For a removed object, say that it is gone,
unless a new object replaces it. For a new or removed relation, describe the resulting
arrangement (\eg, ``the puddle lies on the tray'').

\textbf{Step 3: Keep the rest.}
Do not mention objects that no edit names. End with ``Keep everything else in the image
unchanged.''

\textbf{Output Format \& Example:} \\
Return a structured JSON object exactly matching this schema:
\begin{lstlisting}[language=json, backgroundcolor=\color{gray!10}, xleftmargin=1em]
{
  "instruction": "Replace the ice cube with a puddle of water that spreads over most of the tray, and make the tray wet around it. Keep everything else in the image unchanged."
}
\end{lstlisting}
\end{promptbox}

The instruction $c_i$ and the observed frame $I_0$ are passed to the image editor $\Psi$,
which renders the keyframe $I_i$. Object masks are then extracted with Grounded-SAM-2,
prompted with the category of each object in $\mathcal{G}_i$; when several instances are
found, all of them are kept, since they define the keyframe instances $\{Q_l\}$. Depth maps
are predicted with Depth Anything V2 and normalized to $[0, 1]$ per image before computing
depth differences.

\subsection{Guidance Details}
\label{app:guidance}

\parag{Hyperparameters.}
\label{app:hyperparameters}
GPSR uses Gemini 3 Flash with temperature $0$ and JSON-constrained outputs, and keyframes
are rendered with Gemini 3 Pro Image (Nano Banana Pro). The VDM is CogVideoX-I2V-5B with 50
denoising steps; videos have 49 frames at $720 \times 480$ and are generated with a fixed seed
on one NVIDIA H100 GPU. Guidance follows the CogVideoX defaults of Frame Guidance:
$t_E = 5$, $t_L = 20$, $N_{\mathrm{g}} = 10$, and $\eta = 3.0$.

\parag{Latent slicing and flow matching.}
CogVideoX encodes the first frame separately and compresses the remaining frames by a
factor of $r = 4$ in time, so frame $f$ corresponds to latent index $j(f) = 0$ for $f = 0$
and $j(f) = 1 + \lfloor (f-1)/r \rfloor$ otherwise; the 49 frames map to 13 latents.
Following Frame Guidance, the preview $\hat{x}_i$ is decoded from a window of three latents
around $j(f_i)$, and the frame at $f_i$ is taken from the decoded clip. CogVideoX is a $v$-prediction model, so the clean latent is estimated with Tweedie's
formula as in the main paper. For a VDM trained with rectified flow,
$z_t = (1-t)\, z_0 + t\, \varepsilon$, the estimate becomes
$z_{0|t} = z_t - t\, v_\theta(z_t, t)$, and re-noising follows the flow-matching variant of
the time-travel step in Frame Guidance.

\parag{Occupancy, instances, and regions.}
The encoder $\mathcal{F}$ is DINOv3, which produces a feature map at the stride of its patch size. The preview and the
keyframe are resized to the same resolution before encoding, and masks are resized to the
feature grid by average pooling. Instances in the preview are the connected components of
$\Phi_o(\hat{x}_i) > 0.5$. Centroids are computed in coordinates normalized to $[0, 1]$, so the
matching cost and $\mathcal{L}_{\mathrm{pos}}$ do not depend on resolution.

The region $\Omega_k$ of each term is computed at the image resolution as a binary map and
average-pooled to the latent grid, which gives soft values at its boundary. For
$\mathcal{L}_{\mathrm{depth}}$ on a pair $(a, b)$, $\Omega_k$ is the union of the regions of
$a$ and $b$. For $\mathcal{L}_{\mathrm{count}}$ of an object whose instances are all
matched, the region is still built from the matched pairs, so that the term keeps the
object in place. The same spatial map is applied to every temporal index of the latent.

\parag{Guidance algorithm.}
\Cref{alg:app_guidance} summarizes one sampling run. It follows the schedule of Frame
Guidance: in the first $n_E$ denoising steps, while the layout forms, the latent is updated
deterministically with the region-weighted gradient; in the following steps up to $n_L$, the
unweighted gradient is applied with time-travel, \ie, the guided latent is denoised by one
step and re-noised back to the current noise level, with the number of repeats decreasing
linearly; the remaining steps are unguided. Keyframes, masks, and depth maps are computed
once before sampling.

\begin{algorithm}[!h]
\caption{Graph-Guided Test-Time Optimization (one sampling run)}
\label{alg:app_guidance}
\begin{algorithmic}[1]
\Require net edits and anchors $\{(\bar{\epsilon}_i, f_i)\}_{i=1}^{K}$; keyframes
    $\{I_i\}$ with masks and depth maps; timesteps $t_1 > \dots > t_T$; stage boundaries
    $n_E < n_L$; repeats $N_{\mathrm{g}}$; step size $\eta$; region weight $\lambda$
\State $z_{t_1} \sim \mathcal{N}(0, \mathbf{I})$
\State $\mathcal{J}_i \gets$ latent window covering frame $f_i$, for $i = 1, \dots, K$
    \Comment{latent slicing}
\For{$n = 1, \dots, T$}
    \State $t \gets t_n$
    \If{$n \le n_L$} \Comment{guided steps}
        \State $M \gets N_{\mathrm{g}}$ \textbf{if} $n \le n_E$ \textbf{else}
            $\lceil N_{\mathrm{g}}\,(n_L - n + 1)/(n_L - n_E) \rceil$
            \Comment{repeats decrease linearly}
        \For{$m = 1, \dots, M$}
            \State $z_{0|t} \gets \sqrt{\bar\alpha_t}\, z_t - \sqrt{1 - \bar\alpha_t}\,
                v_\theta(z_t, t)$ \Comment{\cref{eq:tweedie}}
            \State $g \gets 0$
            \For{$i = 1, \dots, K$}
                \State $\hat{x}_i \gets \mathcal{D}\bigl(z_{0|t}[\mathcal{J}_i]\bigr)$
                    \Comment{preview at anchor $f_i$}
                \State compute $\Phi_o(\hat{x}_i)$, instances $\{P_j\}$, and matching
                    $\pi^{*}$ for every object $o$
                \For{each term $\mathcal{L}^{(i)}_k$ of \cref{eq:objective}}
                    \State $\hat{g}_k \gets \nabla_{z_t}\mathcal{L}^{(i)}_k \,/\,
                        \max\bigl(\lVert \nabla_{z_t}\mathcal{L}^{(i)}_k \rVert,
                        \varepsilon\bigr)$ \Comment{normalized gradient}
                    \If{$n \le n_E$}
                        \State $g \gets g + \bigl(\Omega_k + \lambda (1 - \Omega_k)\bigr)
                            \odot \hat{g}_k$ \Comment{region-weighted}
                    \Else
                        \State $g \gets g + \hat{g}_k$
                    \EndIf
                \EndFor
            \EndFor
            \If{$n \le n_E$} \Comment{layout stage: deterministic update}
                \State $z_t \gets z_t - \eta\, g$
            \Else \Comment{detail stage: time-travel}
                \State $z_{t_{n+1}} \gets \mathrm{DDIM}(z_t - \eta\, g,\; t \to t_{n+1})$
                \State $z_t \gets \sqrt{\bar\alpha_t / \bar\alpha_{t_{n+1}}}\; z_{t_{n+1}}
                    + \sqrt{1 - \bar\alpha_t / \bar\alpha_{t_{n+1}}}\;\xi,
                    \quad \xi \sim \mathcal{N}(0, \mathbf{I})$ \Comment{re-noise}
            \EndIf
        \EndFor
    \EndIf
    \State $z_{t_{n+1}} \gets \mathrm{DDIM}(z_t,\; t \to t_{n+1})$ \Comment{reverse step}
\EndFor
\State \Return $\mathcal{D}(z_{t_{T+1}})$ \Comment{$t_{T+1} = 0$}
\end{algorithmic}
\end{algorithm}

\subsection{Formulation of the Guidance Loss}
\label{app:loss}

Each term of \cref{sec:stage2} compares an object-level summary of the preview $\hat{x}_i$ with the
same summary of the keyframe $I_i$, so it is zero when the selected property matches,
whatever contour or texture the editor drew. Every term reaches the latent through the
occupancy or the features of the preview, so we analyze its gradient at a grid location $u$;
the chain rule carries it to $z_t$. We write $\Phi(u) = \Phi_o(\hat{x}_i)(u)$ and
$\Phi^{*}(u) = \Phi_o(I_i)(u)$.

\parag{Soft occupancy.}
With $s(u) = \cos(\mathcal{F}(\hat{x}_i)(u), \rho_o)$,
\begin{equation}
    \Phi(u) = \sigma\Bigl(\frac{s(u) - b}{\tau}\Bigr),
    \qquad
    \frac{\partial \Phi(u)}{\partial s(u)} = \frac{\Phi(u)\bigl(1 - \Phi(u)\bigr)}{\tau}
    \le \frac{1}{4\tau}.
    \label{eq:app_occ}
\end{equation}
The gradient peaks where $s(u) = b$, \ie, at the boundary of the object, and vanishes inside
it and in the background, so all terms below change the object where it can grow, shrink, or
move. A small $\tau$ sharpens the mask but saturates the sigmoid.

\parag{Count term.}
For an unmatched preview instance $P_j$ and an unmatched keyframe instance $Q_l$, with
$r_l = \operatorname{avg}_{Q_l}\Phi / \operatorname{avg}_{Q_l}\Phi^{*}$,
\begin{equation}
    \frac{\partial}{\partial \Phi(u)} \operatorname{avg}_{P_j}\Phi = \frac{1}{|P_j|} > 0,
    \qquad
    \frac{\partial}{\partial \Phi(u)} \max(0, 1 - r_l) =
    -\frac{\mathbf{1}[r_l < 1]}{|Q_l|\,\operatorname{avg}_{Q_l}\Phi^{*}} \le 0.
    \label{eq:app_count}
\end{equation}
Descent lowers the occupancy of an extra copy uniformly until it disappears, and raises it
where a missing copy should be. The fill target is the keyframe's own occupancy, which a
correct object reaches, not one, and the hinge stops the term once it is reached.

\parag{Appearance term.}
With the mean features $p = \bar{\mathcal{F}}_{P_j}(\hat{x}_i)$ and $q =
\bar{\mathcal{F}}_{Q_l}(I_i)$,
\begin{equation}
    \mathcal{L}_{\mathrm{app}} = 1 - \hat{p}^{\top}\hat{q},
    \qquad
    \frac{\partial \mathcal{L}_{\mathrm{app}}}{\partial \mathcal{F}(\hat{x}_i)(u)} =
    -\frac{\hat{q} - (\hat{p}^{\top}\hat{q})\,\hat{p}}{|P_j|\,\lVert p \rVert}
    \quad (u \in P_j).
    \label{eq:app_app}
\end{equation}
The gradient is orthogonal to $\hat{p}$, so it rotates the appearance toward the keyframe
without changing the feature norm, and it is identical at every $u \in P_j$, so it changes
the object as a whole, independently of its shape.

\parag{Extent term.}
With the soft areas $A = \sum_{R_{jl}} \Phi$ and $A^{*} = \sum_{R_{jl}} \Phi^{*}$,
\begin{equation}
\begin{aligned}
    \mathcal{L}_{\mathrm{ext}} &= \log^{2}\frac{A}{A^{*}},
    &\qquad
    \frac{\partial \mathcal{L}_{\mathrm{ext}}}{\partial \Phi(u)} &=
    \frac{2}{A}\log\frac{A}{A^{*}}, \\
    \mathcal{L}_{\mathrm{ext}}(\kappa A^{*}, A^{*}) &=
    \mathcal{L}_{\mathrm{ext}}(A^{*}/\kappa, A^{*}),
    &\qquad
    \mathcal{L}_{\mathrm{ext}}(cA, cA^{*}) &= \mathcal{L}_{\mathrm{ext}}(A, A^{*}).
\end{aligned}
\label{eq:app_ext}
\end{equation}
The sign says whether to grow or shrink; growth and shrinkage by the same factor cost the
same, and small and large objects are treated alike. Summing over $R_{jl} = P_j \cup Q_l$
gives a gradient where the object should grow, which $P_j$ alone would not.

\parag{Position term.}
With the centroid $c = \sum_{R_{jl}} u\,\Phi(u) / A$ and its keyframe counterpart $c^{*}$,
\begin{equation}
    \frac{\partial c}{\partial \Phi(u)} = \frac{u - c}{A},
    \qquad
    \frac{\partial \lVert c - c^{*} \rVert^{2}}{\partial \Phi(u)} =
    \frac{2}{A}\,(c - c^{*})^{\top}(u - c).
    \label{eq:app_pos}
\end{equation}
The gradient is negative on the side of the object facing $c^{*}$ and positive on the other
side, so descent moves the object toward the target instead of deforming it. The region must
therefore contain the target, which $R_{jl}$ and the hull in $\Omega_k$ ensure. Matching the
arrangement of the keyframe realizes a relation that has no differentiable predicate.

\parag{Depth term.}
With the occupancy-weighted depth $D_a = \sum_u \Phi_a(u)\,z(u) / \sum_u \Phi_a(u)$ and
$\Delta z_{ab} = D_a - D_b$,
\begin{equation}
    z \mapsto \alpha z + \beta
    \;\Longrightarrow\;
    D_a \mapsto \alpha D_a + \beta,
    \qquad
    \Delta z_{ab} \mapsto \alpha\,\Delta z_{ab}.
    \label{eq:app_depth}
\end{equation}
The gap is free of the unknown offset of monocular depth, and per-image normalization to
$[0, 1]$ fixes $\alpha$, so the term compares only which object is in front and by how much.

\parag{Combination.}
The terms have different ranges ($\mathcal{L}_{\mathrm{app}} \in [0, 2]$, while
$\mathcal{L}_{\mathrm{ext}}$ is unbounded). The update uses
\begin{equation}
    \hat{g}_k = \frac{\nabla_{z_t}\mathcal{L}_k}{\lVert \nabla_{z_t}\mathcal{L}_k \rVert},
    \qquad
    \hat{g}_k(c\,\mathcal{L}_k) = \hat{g}_k(\mathcal{L}_k) \quad \forall c > 0,
    \label{eq:app_norm}
\end{equation}
so every term contributes a unit direction regardless of its scale, and no weights are
needed. A small constant is added to all denominators, and terms with empty instance sets are
skipped.

\subsection{Example Event Chain}
\label{app:example}

\Cref{tab:app_example} shows the complete event chain for ``an ice cube melting in the
sun'' on a tray, for $F = 49$ frames. At the final state, the net edits are
\opname{Consume}(ice\#1), \opname{Spawn}(puddle\#3 $\leftarrow$ ice\#1),
\opname{Update}(puddle\#3, extent, spread over most of the tray),
\opname{Link}(tray\#2, support, puddle\#3), and \opname{Update}(tray\#2, surface, wet).
The table on which the tray stands (table\#4) is never named by a delta and stays unchanged
throughout.

\begin{table}[!h]
    \centering
    \caption{\textbf{Event chain for ``an ice cube melting in the sun''} ($F = 49$).}
    \label{tab:app_example}
    \small
    \begin{tabularx}{\linewidth}{@{}c>{\raggedright\arraybackslash}X>{\raggedright\arraybackslash}X>{\raggedright\arraybackslash}Xcc@{}}
        \toprule
        \rowcolor{gray!20} $i$ & \textbf{State} $s_o$ & \textbf{Physical rule} $r_o$ &
            \textbf{Edits} $\epsilon_i$ & $d_i$ & $f_i$ \\
        \midrule
        1 & ice\#1: partially melted; tray\#2: wet beneath ice\#1
          & ice above its melting point turns into water; liquid flows down onto its
            support
          & \opname{Update}(ice\#1, material phase, partially melted),
            \opname{Update}(tray\#2, surface, wet)
          & 0.30 & 14 \\
        2 & ice\#1: melted into puddle\#3 on tray\#2
          & ice above its melting point turns into water
          & \opname{Spawn}(puddle\#3 $\leftarrow$ ice\#1), \opname{Consume}(ice\#1),
            \opname{Link}(tray\#2, support, puddle\#3)
          & 0.30 & 29 \\
        3 & puddle\#3: spread over most of tray\#2
          & liquid spreads over a flat support
          & \opname{Update}(puddle\#3, extent, spread over most of the tray)
          & 0.25 & 41 \\
        \bottomrule
    \end{tabularx}
\end{table}

\section{Additional Quantitative Results}
\label{app:full_results}

\Cref{tab:vbench} reports the
quality dimensions of VBench, including closed-source VDMs.

\begin{table}[t!]
    \centering
    \caption{Quality dimensions of VBench \citep{huang2024vbench}. QS: Quality Score; SC:
    Subject Consistency; BC: Background Consistency; TF: Temporal Flickering; MS: Motion
    Smoothness; DD: Dynamic Degree; AQ: Aesthetic Quality; IQ: Imaging Quality. Best results
    are in \textbf{bold}.}
    \label{tab:vbench}
    \begin{tabular}{l c c c c c c c c}
        \toprule
        \textbf{Method} & \textbf{QS}$\uparrow$ & \textbf{SC}$\uparrow$ & \textbf{BC}$\uparrow$
            & \textbf{TF}$\uparrow$ & \textbf{MS}$\uparrow$ & \textbf{DD}$\uparrow$
            & \textbf{AQ}$\uparrow$ & \textbf{IQ}$\uparrow$ \\
        \midrule
        \multicolumn{9}{c}{\textit{Closed-source VDMs}} \\
        \midrule
        Runway Gen-3 & 84.11 & 97.10 & 96.62 & 98.61 & 99.23 & 60.14 & 63.34 & \textbf{66.82} \\
        Kling        & 83.39 & \textbf{98.33} & \textbf{97.60} & 99.30 & 99.40 & 46.94 & 61.21 & 65.62 \\
        Pika         & 82.92 & 96.94 & 97.36 & \textbf{99.74} & \textbf{99.50} & 47.50 & 62.04 & 61.87 \\
        Luma         & 83.47 & 97.33 & 97.43 & 98.64 & 99.35 & 44.26 & \textbf{65.51} & 66.55 \\
        \midrule
        \multicolumn{9}{c}{\textit{Open-source}} \\
        \midrule
        CogVideoX-I2V-5B & 83.05 & 96.45 & 96.71 & 98.97 & 97.20 & 69.51 & 61.88 & 63.33 \\
        \rowcolor{oursrow}
        \textbf{PhysPlan (Ours)} & \textbf{84.88} & 97.06 & 97.10 & 98.72 & 98.80
            & \textbf{75.62} & 62.20 & 65.78 \\
        \bottomrule
    \end{tabular}%
    \end{table}

\section{Failure Analysis}
\label{app:failure}

We analyze where PhysPlan fails on Physics-IQ. A video counts as failed if its score is in
the lowest third of all PhysPlan videos, which gives 66 failed videos. Two authors inspected
each of them, together with its event chain and keyframes, and attributed it to the first
stage of the pipeline that goes wrong:
\begin{itemize}[leftmargin=*]
    \item \textbf{Reasoning}: the deltas are physically wrong or incomplete, \eg, a missing
    consequence on a passive object or a wrong order of events.
    \item \textbf{Edits}: the edit set is rejected by the checks after all retries, or
    passes them but misrepresents the delta.
    \item \textbf{Keyframe}: the image editor does not realize an edit, or changes objects
    that no edit names.
    \item \textbf{Localization}: masks, occupancy, or depth are wrong, \eg, for transparent,
    reflective, thin, or fragmented objects.
    \item \textbf{Guidance}: the event chain and keyframes are correct, but the video does not
    follow them, typically for continuous motion between anchors.
\end{itemize}
Disagreements between the two authors were resolved by discussion.

\begin{table}[!h]
    \centering
    \caption{\textbf{Failure attribution} on Physics-IQ: share of failed videos (\%) per
    pipeline stage and category. Each row sums to 100.}
    \label{tab:app_failure}
    \small
    \begin{tabular}{@{}lccccc@{}}
        \toprule
        \rowcolor{gray!20} \textbf{Category} & \textbf{Reasoning} & \textbf{Edits} &
            \textbf{Keyframe} & \textbf{Localization} & \textbf{Guidance} \\
        \midrule
        Solid Mechanics & 12 & 4 & \textbf{30} & 26 & 28 \\
        Fluid Dynamics  & 8  & 3 & 20 & 29 & \textbf{40} \\
        Optics          & 10 & 2 & 14 & \textbf{52} & 22 \\
        Magnetism       & 18 & 5 & 12 & \textbf{35} & 30 \\
        Thermodynamics  & 21 & 6 & \textbf{30} & 18 & 25 \\
        \midrule
        \textbf{All}    & 11 & 4 & 25 & \textbf{30} & \textbf{30} \\
        \bottomrule
    \end{tabular}
\end{table}

\begin{table}[!h]
    \centering
    \caption{\textbf{Deterministic checks} over all edit sets: share that pass at the first
    attempt, after retries, or not at all, and share of each check among all violations.}
    \label{tab:app_checks}
    \small
    \begin{tabular}{@{}lc@{}}
        \toprule
        \rowcolor{gray!20} \textbf{Statistic} & \textbf{Value (\%)} \\
        \midrule
        Pass at first attempt          & 87 \\
        Pass after retries             & 97 \\
        Rejected after all retries     & 3 \\
        \midrule
        Violations: coverage           & 46 \\
        Violations: grounding          & 31 \\
        Violations: consistency        & 15 \\
        Violations: lineage            & 8 \\
        \bottomrule
    \end{tabular}
\end{table}

\parag{Discussion.}
Most failures happen after the plan is made (\cref{tab:app_failure}), and the dominant cause
depends on the category. Guidance dominates only Fluid Dynamics (40\%): the event chain fixes
the state of the scene only at the anchors, and continuous flow between them is left to the
VDM, which is consistent with the gap to VLIPP in this category. Localization dominates
Optics (52\%) and Magnetism (35\%), whose scenes often contain transparent, reflective, or
small metallic objects, for which feature similarity and monocular depth are least reliable.
Keyframe failures dominate Solid Mechanics and Thermodynamics (30\% each): in Solid
Mechanics, the editor often draws an implausible pose or arrangement after a fall or a
collision, and in Thermodynamics, it tends to miss subtle changes such as a slight change of
color or phase. Overall, localization and guidance each account for 30\% of the failures and
keyframes for 25\%. Reasoning errors are less frequent (11\%) and concentrate in
Thermodynamics and Magnetism, whose phenomena are less common and whose effects on passive
objects are harder to predict. Failures of the edits are rare (4\%): as \cref{tab:app_checks}
shows, 87\% of the edit sets pass the checks at the first attempt and 97\% after retries.
Coverage is the most frequently violated check, \ie, the VLM most often edits an object that
the delta does not name or forgets one that it does; this is exactly the error that the
checks prevent from reaching the state graph.

\section{Cost Analysis}
\label{app:cost}

\Cref{tab:app_cost_stage} breaks down the average runtime of PhysPlan per video,
\cref{tab:app_cost_api} estimates its API cost, and \cref{tab:app_cost_compare} compares its
total cost with the base model and Frame Guidance. All measurements use a single NVIDIA H100
(80GB) GPU, CogVideoX-I2V-5B, and $720 \times 480$ videos of 49 frames, with $K = 3$ events
unless stated otherwise.

\begin{table}[!h]
    \centering
    \caption{\textbf{Runtime per stage} of PhysPlan (seconds per video, $K = 3$).}
    \label{tab:app_cost_stage}
    \small
    \begin{tabular}{@{}llc@{}}
        \toprule
        \rowcolor{gray!20} \textbf{Stage} & \textbf{Model} & \textbf{Time (s)} \\
        \midrule
        \multicolumn{3}{@{}l}{\textit{Grounded Physical State Reasoning}} \\
        \quad Scene parsing and decomposition & Gemini 3 Flash & 6.8 \\
        \quad Delta translation and checks    & Gemini 3 Flash & 4.2 \\
        \quad Keyframe rendering              & Gemini 3 Pro Image & 12.5 \\
        \quad Masks and depth                 & Grounded-SAM-2, Depth Anything V2 & 4.0 \\
        \midrule
        \multicolumn{3}{@{}l}{\textit{Graph-Guided Test-Time Optimization}} \\
        \quad Preview decoding                & CogVideoX VAE & 24.5 \\
        \quad Measurement and backpropagation & DINOv3, CogVideoX & 214.0 \\
        \quad Sampling and final decoding     & CogVideoX & 100.0 \\
        \midrule
        \textbf{Total}                        & & 366.0 \\
        \bottomrule
    \end{tabular}
\end{table}

\parag{API cost.}
We estimate the API cost from the average number of tokens per call and the standard list
prices of the Gemini API at the time of writing: \$0.50 and \$3.00 per million input and
output tokens for Gemini 3 Flash, whose output tokens include its thinking tokens, and
\$0.134 per output image at 1K resolution for Gemini 3 Pro Image. Delta translation includes
the retries triggered by the checks. \Cref{tab:app_cost_api} lists the estimate for $K = 3$.

\begin{table}[!h]
    \centering
    \caption{\textbf{Estimated API cost} of PhysPlan per video ($K = 3$), at standard Gemini
    API list prices.}
    \label{tab:app_cost_api}
    \small
    \begin{tabular}{@{}lccc@{}}
        \toprule
        \rowcolor{gray!20} \textbf{Call} & \textbf{Calls} & \textbf{Tokens (in / out)} &
            \textbf{Cost (USD)} \\
        \midrule
        Scene parsing ($p_{\mathrm{parse}}$)          & 1   & 2.5k / 2.5k & 0.009 \\
        Phenomenon decomposition ($p_{\mathrm{delta}}$) & 1 & 3.5k / 3.0k & 0.011 \\
        Delta translation ($p_{\mathrm{edit}}$)       & 3.4 & 7.5k / 2.7k & 0.012 \\
        Keyframe instruction ($p_{\mathrm{render}}$)  & 3   & 7.8k / 1.2k & 0.008 \\
        Keyframe rendering (Gemini 3 Pro Image)       & 3   & 3 images    & 0.405 \\
        \midrule
        \textbf{Total}                                & & & 0.445 \\
        \bottomrule
    \end{tabular}
\end{table}

\begin{table}[!h]
    \centering
    \caption{\textbf{Total cost} per video, and its growth with the number of events $K$.
    Frame Guidance takes keyframes as input and therefore has no API cost of its own.}
    \label{tab:app_cost_compare}
    \small
    \begin{tabular}{@{}lccc@{}}
        \toprule
        \rowcolor{gray!20} \textbf{Method} & \textbf{Time (s)} & \textbf{Peak memory (GB)} &
            \textbf{API cost (USD)} \\
        \midrule
        CogVideoX-I2V-5B          & 100 & 26 & --- \\
        Frame Guidance            & 310 & 58 & --- \\
        PhysPlan                  & 366 & 64 & 0.45 \\
        \midrule
        PhysPlan, $K = 3$         & 366 & 64 & 0.45 \\
        PhysPlan, $K = 5$         & 471 & 70 & 0.73 \\
        PhysPlan, $K = 7$         & 575 & 76 & 1.01 \\
        \bottomrule
    \end{tabular}
\end{table}

\parag{Discussion.}
Reasoning about the scene is cheap in time: all GPSR steps, including keyframe rendering and
grounding, take 27.5\,s, or 7.5\% of the total (\cref{tab:app_cost_stage}), and run once per
video. Its API cost is about \$0.45 per video (\cref{tab:app_cost_api}), of which over 90\%
comes from rendering the keyframes; all VLM calls together cost about \$0.04. Evaluating a
full benchmark of 200 videos thus costs about \$90 in API calls, and the Batch API of Gemini,
which halves all prices, would reduce this further. The runtime is dominated by test-time
optimization, in particular by backpropagating through the denoiser at every guided step,
which is shared with Frame Guidance. Compared with Frame Guidance, PhysPlan adds 56\,s (18\%)
and 6\,GB of memory (\cref{tab:app_cost_compare}), mainly for the feature encoder, the
instance matching, and the additional terms. Runtime and API cost both grow roughly linearly
with the number of events, by about 52\,s and \$0.14 per event, since every event adds one
keyframe to render and one anchor whose preview is decoded and whose terms enter the
objective. Compared with training-based methods, which require fine-tuning the VDM on curated
data, this cost is paid only at inference and requires no training.

\section{User Study}
\label{app:user_study}

\begin{figure}[!h]
    \centering
    \includegraphics[width=\linewidth]{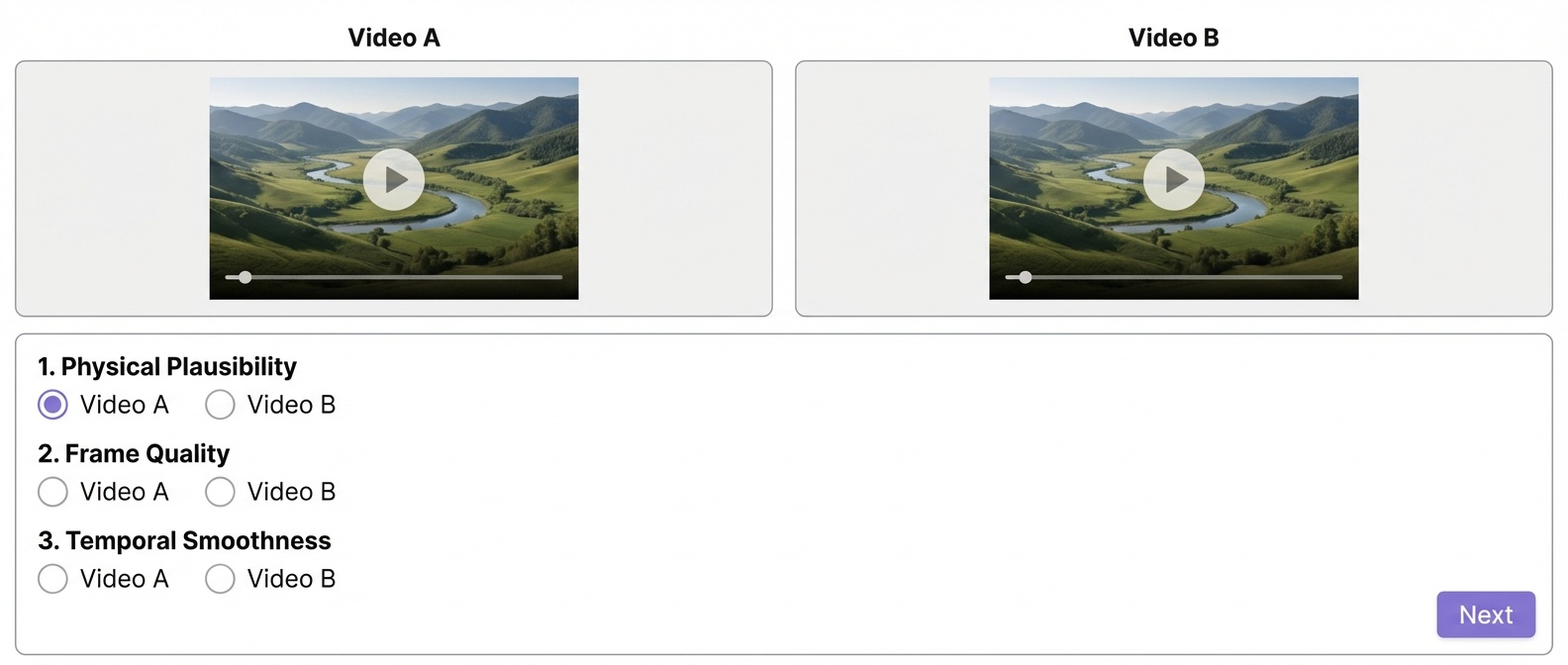}
    \caption{\textbf{User study interface.} Each trial shows two videos from the same prompt
    and initial frame, one by PhysPlan and one by a baseline, in random order. The participant
    chooses the better video for each of the three criteria and continues with \emph{Next}.}
    \label{fig:app_us_interface}
\end{figure}

\parag{Protocol.}
We conduct a two-alternative forced-choice (2AFC) study with the web interface shown in
\cref{fig:app_us_interface}. In each trial, a participant sees two videos, labeled Video A
and Video B, generated from the same prompt and the same initial frame: one by PhysPlan and
one by a baseline, either CogVideoX-I2V-5B or Frame Guidance. The two videos are shown side by
side and loop in sync, so that the same moment of the phenomenon is visible in both at the
same time. For each of three criteria, the participant chooses the better video:
\begin{itemize}[leftmargin=*]
    \item \textbf{Physical plausibility}: which video follows the described phenomenon in a
    physically plausible way, including its effects on other objects in the scene?
    \item \textbf{Frame quality}: which video has sharper, cleaner, and more realistic
    frames?
    \item \textbf{Temporal smoothness}: which video changes more smoothly over time, without
    flickering, jumps, or drifting background?
\end{itemize}
A choice is required for every criterion before the participant can move to the next trial,
so there are no ties. The left or right position of each method is randomized in every
trial, and participants are not told which methods are compared or which video comes from
which method. Before the study, participants see a short description of the three criteria
with one example trial, which is not counted.

\parag{Prompts and workload.}
We sample 40 prompts, 20 from PhyGenBench and 20 from Physics-IQ, covering all their domains,
and generate one pair per prompt and baseline, which gives 80 pairs. Each participant rates
40 randomly assigned pairs, so that every pair is rated by 30 participants on average and the
study collects 2{,}400 ratings per criterion. \Cref{tab:app_us_setup} summarizes the setup.

\begin{table}[!h]
    \centering
    \caption{\textbf{Setup of the user study.}}
    \label{tab:app_us_setup}
    \small
    \begin{tabular}{@{}lc@{}}
        \toprule
        \rowcolor{gray!20} \textbf{Item} & \textbf{Value} \\
        \midrule
        Participants                          & 60 \\
        Prompts (PhyGenBench / Physics-IQ)    & 20 / 20 \\
        Baselines                             & CogVideoX-I2V-5B, Frame Guidance \\
        Video pairs                           & 80 \\
        Pairs rated per participant           & 40 \\
        Ratings per pair (average)            & 30 \\
        Ratings per criterion (total)         & 2{,}400 \\
        \bottomrule
    \end{tabular}
\end{table}

\parag{Preference rate.}
For a criterion $c$ and a baseline $B$, let $V_c(B)$ be the set of all votes cast on pairs
that compare PhysPlan with $B$. The preference rate for PhysPlan is the share of these votes
that choose PhysPlan,
\begin{equation}
    \mathrm{Pref}_c(B) = \frac{\bigl|\{v \in V_c(B) : v = \text{PhysPlan}\}\bigr|}
    {|V_c(B)|}.
    \label{eq:app_pref}
\end{equation}
Since every vote chooses one of the two videos, the preference rate for the baseline is
$1 - \mathrm{Pref}_c(B)$, and $50\%$ means no preference. The pooled rate reported in the main
paper uses the votes against both baselines together; since both baselines receive the same
number of votes, it is the mean of the two per-baseline rates. \Cref{tab:app_us_results}
reports the rates.

\begin{table}[!h]
    \centering
    \caption{\textbf{Preference rates} (\%) for PhysPlan, per baseline and pooled.}
    \label{tab:app_us_results}
    \small
    \begin{tabular}{@{}lccc@{}}
        \toprule
        \rowcolor{gray!20} \textbf{Against} & \textbf{Physical plausibility} &
            \textbf{Frame quality} & \textbf{Temporal smoothness} \\
        \midrule
        CogVideoX-I2V-5B & 76 & 63 & 77 \\
        Frame Guidance   & 68 & 57 & 69 \\
        \midrule
        \textbf{Pooled}  & 72 & 60 & 73 \\
        \bottomrule
    \end{tabular}
\end{table}

PhysPlan is preferred against both baselines on all criteria. The margin is smaller against
Frame Guidance, which uses the same keyframes, and smallest for frame quality, where Frame
Guidance is competitive, consistent with its similar FID on PhyGenBench (46.4 vs.\ 45.4). The
margins in physical plausibility and temporal smoothness remain clear, matching the gains on
the benchmarks and in FVD.

\parag{Participants.}
The study involved 60 adult volunteers, aged 19 to 38 (median 24), recruited from our
university through mailing lists and student groups. About two thirds (38) have a background
in computer science or engineering, 12 in the natural sciences, and 10 in other fields such as
design and the humanities. Most participants are familiar with AI-generated video but do not
work with it: 21 see generated videos regularly, 27 occasionally, and 12 rarely or never. No
participant was involved in this work. Participation was voluntary and anonymous: all
participants gave informed consent, no personally identifiable information was collected,
and the study was granted an exemption by the institutional review board.